\documentclass{tut2016} 

\usepackage{amsfonts,bm}
\usepackage{subcaption, graphicx}

\usepackage[colorlinks=true,breaklinks=true,bookmarks=false,urlcolor=blue,%
  citecolor=blue,linkcolor=blue,bookmarksopen=false,draft=false]{hyperref}
\def\R{\mathbb{R}}

\begin{document}

\CHAPTERNO{}
\TITLE{The Secrets of Machine Learning: Ten Things You Wish You Had Known Earlier to be More Effective at Data Analysis}    

\AUBLOCK{%
  \AUTHOR{Cynthia Rudin}
  \AFF{Computer Science, Electrical and Computer Engineering, and Statistical Science, Duke University, 
       \EMAIL{cynthia@cs.duke.edu}} 
  \AUTHOR{David Carlson}
  \AFF{Civil and Environmental Engineering, Biostatistics and Bioinformatics, Electrical and Computer Engineering, and Computer Science, Duke University,
       \EMAIL{david.carlson@duke.edu}} 
} 

\CHAPTERHEAD{The Secrets of Machine Learning}

\ABSTRACT{Despite the widespread usage of machine learning throughout organizations, there are some key principles that are commonly missed. In particular: 1) There are at least four main families for supervised learning: logical modeling methods, linear combination methods, case-based reasoning methods, and iterative summarization methods. 2) For many application domains, almost all machine learning methods perform similarly (with some caveats). Deep learning methods, which are the leading technique for computer vision problems, do not maintain an edge over other methods for most problems (and there are reasons why). 3) Neural networks are hard to train and weird stuff often happens when you try to train them. 4) If you don't use an interpretable model, you can make bad mistakes. 5) Explanations can be misleading and you can't trust them. 6) You can pretty much always find an accurate-yet-interpretable model, even for deep neural networks. 7) Special properties such as decision making or robustness must be built in, they don't happen on their own. 8) Causal inference is different than prediction (correlation is not causation). 9) There is a method to the madness of deep neural architectures, but not always. 10) It is a myth that artificial intelligence can do anything. 

%
} 


\maketitle  

\section*{Introduction} 

There is much more to machine learning than the current cloud of myth and hype surrounding it. Used in the right ways, the tools of machine learning can make endless numbers of processes more efficient, government more effective, businesses more profitable, and medical patients more safe, but used in the wrong ways, machine learning tools can cause (and have caused!) serious harm. Succinctly, machine learning involves key principles that are often missed, despite its widespread use. Our goal in this tutorial is to help prevent such serious harm by educating you on how to see past the hype and understand what really makes machine learning successful. After you read this, you may be able to find that you can tell fairly quickly whether someone truly knows machine learning, or whether they simply want you think they know it when they really do not. Let us start with the basics.

\section{There are at least four main families of machine learning models for supervised learning}
Machine learning is a huge field that has been applied to diverse problems. At its core is the problem of (supervised) classification, where the basic goal is to predict the answers to yes or no questions (``Will this medical patient suffer a stroke next year?" or  ``Will this person click on this online advertisement?" or perhaps ``Will this criminal be arrested for another crime next year?'' or ``Did this person say `yes' to the voice assistant's question?"). Supervised classification algorithms read your addresses on mail when you send it through the postal service, they read your checks at the bank, they process your spoken words when you speak into your voice assistant, and they even cancel echoes when you speak over a phone line. They are used to recommend products on the internet, provide search returns, provide credit risk scores, suggest spelling corrections to typed text, and predict pollution levels during wildfires. You may not want them to be used in so many circumstances because of privacy concerns about your data (e.g., ``Should companies know my search history and use it to sell me products?"), bias against your gender or ethnic group, or because they cannot be trusted for high-stakes decisions. However, they seem to be here to stay, and despite all their drawbacks, they have been unquestionably useful for many applications that you probably use every day, and applications that you could envision being useful in the future.

Regression is a separate problem from classification, where the goal is to predict a numerical value rather than a yes/no answer (e.g., ``How much will my house sell for if I put it on the market?" or ``How much rain will fall next week?"). Many of the same supervised machine learning techniques developed for classification can be used for regression and vice versa. There are many other kinds of machine learning problems (e.g., density estimation, clustering, data summarization). Our illustrative examples focus mainly on classification and regression.

There are at least four main families of ``supervised'' learning algorithms:  i) logical models (decision trees, rule-based models), ii) linear combinations of trees, stumps or other kinds of features (logistic regression, boosting, random forests, additive models), iii) case-based reasoning, including $K$-nearest neighbors and kernel based methods (support vector machines with gaussian kernels, kernel regression), and iv) iterative summarization (neural networks).

\begin{figure}[t]
    \centering
    \includegraphics[width=\textwidth]{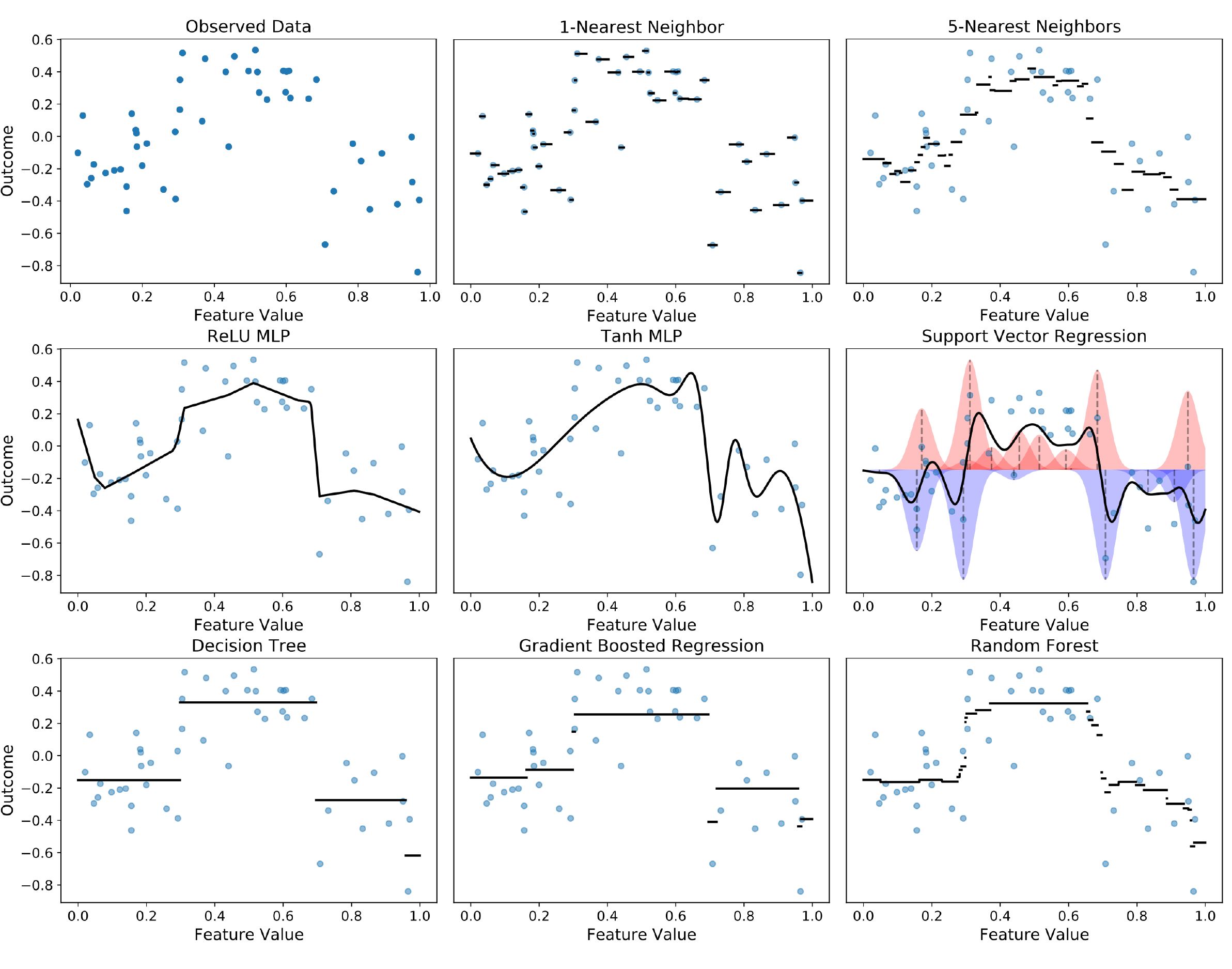}
    \caption{A visual comparison of many different machine learning algorithms applied to a 1-dimensional regression problem.  The top left figure shows the original observed data.  In all other figures, the solid black line shows the predicted outcome from the machine learning method at each feature value. These methods are representative examples of our four main families of supervised machine learning algorithms.  The 1- and 5-Nearest Neighbors and the Kernel Support Vector Regression are examples of case-based reasoning.  The Decision Tree is an example of a logical model.  The Gradient Boosted Regression and the Random Forest are linear combinations or ensembles of  decision trees.  Finally, there are two types of Multi-Layer Perceptions (MLPs), one where a Hypberbolic Tangent (Tanh) non-linearity is used to give smooth transitions, and another where the Rectified Linear Unit (ReLU, $f(x)=\max (x,0)$) is used, which is piecewise linear.  Visually, it can be seen that many of these machine learning methods will give similar prediction quality on these data.}
    \label{fig:alg_comp}
\end{figure}

Figure \ref{fig:alg_comp} illustrates the behavior of these types of models on a regression problem, using a variety of supervised learning methods. There are some observations we can make from looking at the black fitted models within Figure \ref{fig:alg_comp}, without knowing much about the algorithms used for fitting the data. In particular, the parameters of the algorithm can change the model dramatically, as shown from the top two figures on the right, which are both nearest neighbor classifiers. The two figures in the middle row on the left are both multilayer perceptrons, a type of neural network, that produce completely different results by changing one setting.  Also, Figure \ref{fig:alg_comp} shows that many different types of algorithms can perform approximately equally well, depending on the choice of parameters. We will get back to these points later on.

\begin{figure}
    \centering
    \begin{subfigure}[b]{0.45\textwidth}
    \includegraphics[width=0.9\textwidth]{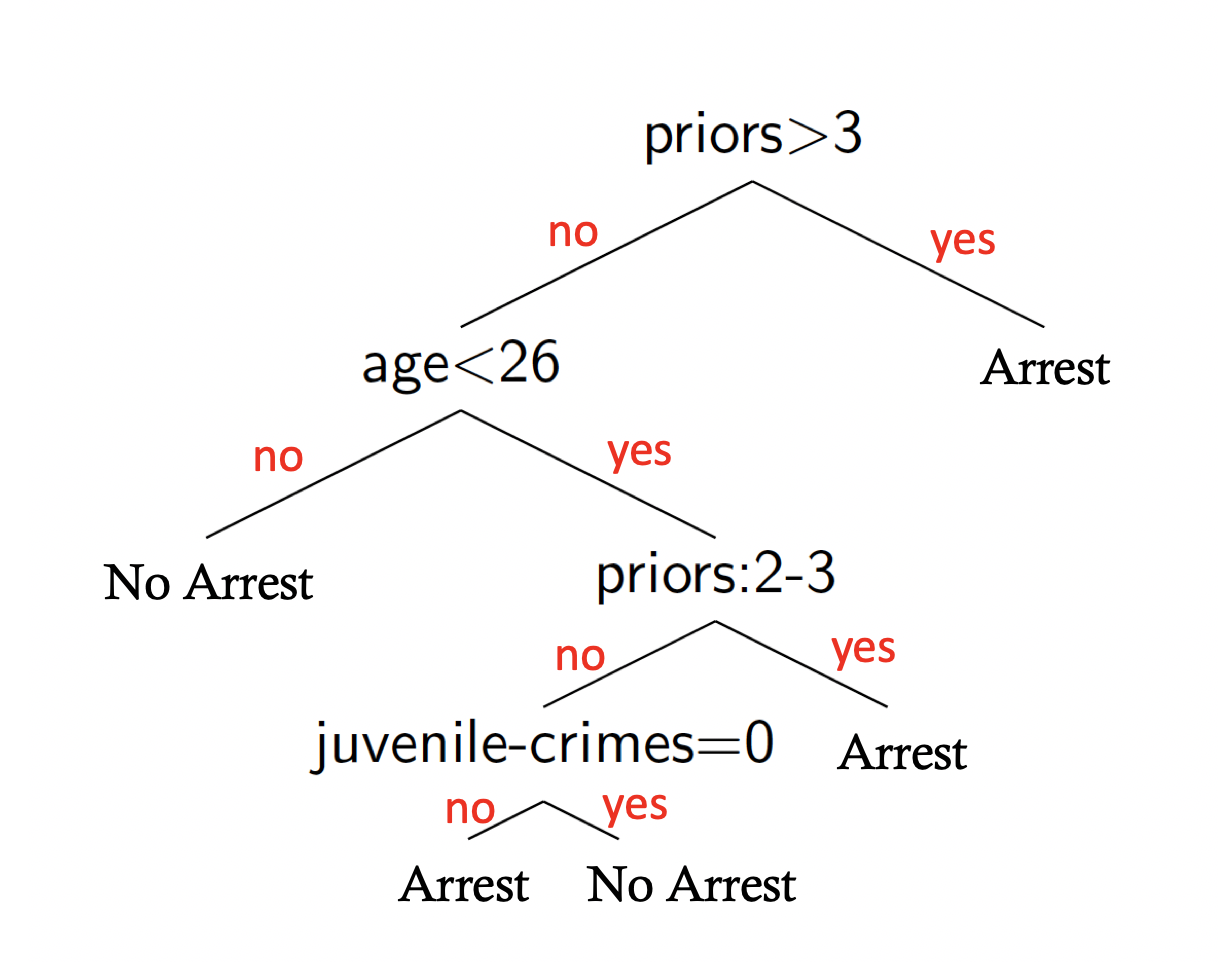}
    \end{subfigure}%
    ~ 
    \begin{subfigure}[b]{0.45\textwidth}
    \begin{small}
    \begin{tabular}{c|c}
        History of stroke &  8 points\\
        Age $<$ 65 years & 0 points\\
        Age 65-74 years & 3 points\\
        Age 74-84 years & 5 points\\
        Age $\geq$ 85 years & 6 points\\
        Female & 1 point \\
        Diabetes Mellitus & 1 point\\
        Past Congestive Heart Failure & 1 point\\
        History of Hypertension & 1 point\\
        Proteinuria & 1 point\\
        eGFR $<$45 or renal disease & 1 point\\
        \hline\\
        Total score & \\
        \hline
    \end{tabular}
    \end{small}
    \vspace*{4pt}
    
    (0-5 points is Low risk)\\ (6 points is Intermediate risk)\\ (7+ points is High risk)\\
    \end{subfigure}
    \caption{The left figure is an example of a decision tree. This decision tree, from \cite{HuRuSe19}, predicts whether someone will be arrested within 2 years, based on their age, number of prior crimes, and whether they have any juvenile criminal history. For someone with more than three prior crimes, the model would predict that the person is likely to be arrested within 2 years. If not, the model would consider whether the person is younger than 26. If the person is 26 or older, the model would predict that the person is not likely to be arrested within 2 years. For the remaining individuals (younger people with 3 or fewer priors), the model then considers number of priors and whether the person had any juvenile crimes to predict future arrest. The right figure, from \cite{ATRIA} is the ATRIA score for predicting stroke in patients with atrial fibrillation. It is a non-linear function of age, and there are no cross-terms involving multiple features. This model is a scoring system, since it is a sparse linear model with integer coefficients.}
    \label{fig:decisiontreeadditive}
\end{figure}

Figure \ref{fig:decisiontreeadditive} illustrates decision trees and additive models for a classification problem. Decision trees are logical models, consisting of if-then rules. An example of a decision tree for predicting criminal recidivism is in Figure \ref{fig:decisiontreeadditive} (left). Small decision trees are interpretable, and possibly mimic the logical thinking of humans. An additive model is in Figure \ref{fig:decisiontreeadditive} (right) for a medical example of predicting stroke in atrial fibrillation patients. An additive model is a sum of non-linear transformations of the original variables. Figure \ref{fig:decisiontreeadditive} shows simpler models, but most of the time, machine learning models are very complicated.

The usual setup is that we are given training data, $\{(x_i,y_i)\}_{i=1}^n$, that are independent and identically distributed (i.i.d.) samples from a probability distribution on a space $\mathcal{X}\times\mathcal{Y}$. Usually, $x_i$ are real-valued covariates $(\mathcal{X}\subset \R^p)$, for instance representing age, gender, and medical history. (Covariates are also called variables or features.) The $y_i$ are labels, such as whether the medical patient suffered a stroke. In reality, training data are almost never i.i.d. from the distribution of interest, but the models built from the training data typically work as long as the true distribution is not too far from the training distribution.

When we use complicated models, we can capture more nuance in the data; however, we risk the possibility that they may overfit the training data, meaning that the model captures random statistical fluctuations rather than real information.  Such models will fail to generalize when applied to new real-world data. Ideally we would choose models that do not overfit, but a model also needs to be complex and flexible enough to fit the data well. Let us discuss this further.

\subsection{Overfitting}
Overfitting is the bane of any model fitting method, and especially in machine learning algorithms. A guiding principle behind any modeling approach is generalization, which is the algorithm's ability to perform well on data it has not seen previously. This is especially an issue in machine learning methods that are ``complex'' in some manner, as they do not generalize to the real world from a small training dataset.
Consider the 1-Nearest Neighbor regression in Figure \ref{fig:alg_comp}, which would perfectly predict every example in the training data (this is like predicting the sale price of a house that has already been sold, and whose price is within the algorithm's database).  It is a much more difficult task to predict out-of-sample, for example on houses that are on the real estate market and have not yet been sold. There are principled mathematical and statistical theories governing generalization, one of which is called \textit{statistical learning theory} \cite{vapnik1998statistical}. In some sense, statistical learning theory is a mathematical way to state Occam's Razor, where Occam's Razor is the idea that among equally accurate explanations, the simpler explanations are better. That is, if we can find a machine learning model that performs well on the training data we have, and yet is somehow ``simple,'' then it is likely to predict well out-of-sample.   In other words, the ``simple'' model will work almost as well on data it has not seen before as on the training set; however, there is a tradeoff because a ``simple'' model may not capture as much nuance of the data as a more complex model. Statistical learning theory mathematically address this tradeoff.

Statistical learning theory is mathematically very beautiful. It states that out-of-sample error (also called test error) is often not too much more than the training error (in-sample error) for a ``simple" model. The most basic learning-theoretic bound is called the Occam's Razor Bound, whose proof is a combination of Hoeffding's inequality and the union bound. 

Let us formally define the Occam's Razor bound. We are given training data $\{(x_i,y_i)\}_{i=1}^n$ drawn independent and identically distributed from a distribution on a space $\mathcal{X}\times\mathcal{Y}$. Usually, each $x_i$ consists of  real-valued covariates $(\mathcal{X}\subset \R^p)$. The $y_i$s are labels. Define $\mathcal{F}$ to be a finite set of models where $f\in\mathcal{F}$ is a map from features to labels, $f:\mathcal{X}\rightarrow\mathcal{Y}$. The total number of possible models is given by the cardinality $|\mathcal{F}|$. An algorithm would pick one model from the set $\mathcal{F}$ to make predictions. A ``simple'' function class would have a small number of possible models (i.e., $|\mathcal{F}|$ is small). This is a simplification compared to the continuous function classes typically used in machine learning algorithms, but is useful to explain the theoretical concepts; the theory can also be extended to continuous spaces.
Let us define the true risk of a function $f$ to be its expected performance over the complete distribution, 
\[
\textrm{TrueRisk}(f):= \mathbb{E}_{x,y}\textrm{loss}(f(x),y),
\]
where the loss is a function giving values between 0 and 1. Here, $\textrm{loss}(f(x),y)$ measures how close our prediction $f(x)$ is to the truth $y$. Poor predictions lead to larger values of the loss. (In reality, if the loss takes values larger than 1, we would scale it to have maximum value 1.) This true risk will give us our performance on out-of-sample data; in practice, we cannot compute the true risk because we have only a finite sample, not the whole distribution, so instead will use a probabilistic bound on the true risk. What we can calculate is the empirical risk of $f$, defined as $f$'s average performance on the training data:
\[
\textrm{EmpRisk}(f):=\frac{1}{n}\sum_{i=1}^n\textrm{loss}(f(x_i),y_i).
\]
If we had overfit, the empirical risk would be very low, and the true risk would be very high. If we did not overfit, the true and empirical risks would be close together. Thus, we would like to know that the true and empirical risk will be close, no matter which model $f$ our algorithm chooses.
\begin{theorem} (\textbf{Occam's Razor Bound})
With probability at least 1-$\delta$ over the random draw of the training data, for all functions $f$ in function class $\mathcal{F}$ of functions mapping $\mathcal{X}$ to $\mathcal{Y}$, the following holds:
\begin{equation}
     \textrm{\rm TrueRisk}(f)\leq   \textrm{\rm EmpRisk}(f) +\sqrt{\frac{\log(|\mathcal{F}|)+\log(1/\delta)}{2n}}.
\end{equation}
\end{theorem}
This theorem tells us that as long as the size of the training set $n$ is large enough, and as long as the size of the function class $|\mathcal{F}|$ is relatively small and not too complex, then our algorithm (no matter which model $f$ it picks) will generalize from the empirical risk (in-sample) to the true risk (out-of-sample). This is called a uniform bound because it holds for all $f$ in $\mathcal{F}$.

\begin{figure}[h]
    \centering
    \includegraphics[width=0.5\textwidth]{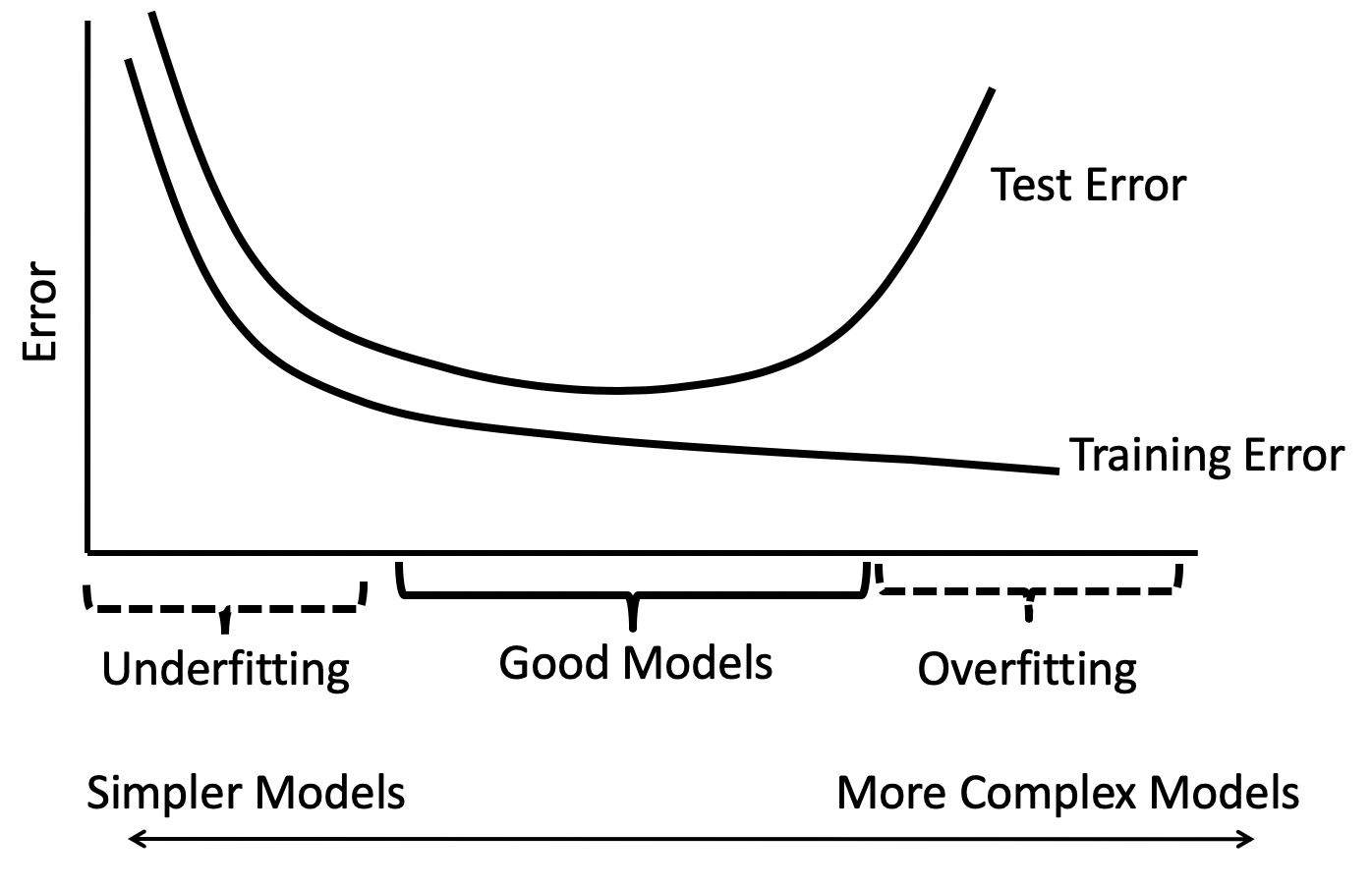}
    \caption{Statistical learning theory's concept of structural risk minimization. For simpler models (towards the left) the training error and test error are close together, but the models do not fit well. For more complex models (towards the right), the training error is small but there is a large gap between training and test performance, and test performance is poor.  The ``good models" are those that balance model complexity with the ability to fit the data. They have the lowest test error and a small gap between training and test error.}  
    \label{fig:srm}
\end{figure}

Here, we have shown a learning-theoretic bound for a finite function class, but in general, the difference between the training and test errors is bounded by a function of the \textit{complexity} of the function class $\mathcal{F}$ that we are using. In other words, when $|\mathcal{F}|$ is infinite, the Occam's razor bound above becomes vacuous. In that case, the term $|\mathcal{F}|$ in the bound can be replaced with a different notion of how complex the function class is, so that the overall bound stays finite.
If we restrict ourselves to linear models in a low-dimensional space, for instance, the complexity of the class of functions is low. If we use highly non-linear models, the complexity of the class could be much higher. Figure \ref{fig:srm} shows the main concept of statistical learning theory, called structural risk minimization. 

Statistical learning theory does not prescribe exactly what types of functions we should use, but it does tell us mathematically how to measure the simplicity (or complexity) of a set of functions. Complexity measures for function classes include various kinds of mathematical covering numbers, Vapnik-Chervonenkis dimension, Rademacher complexity, and other more abstract concepts that we won't discuss here. For instance, if we are able to fit random noise very well with our class of functions, this class of functions is too complex and we are likely to overfit the data. We can keep the models simple by encouraging them to be smooth, adding various kinds of bias, limiting the number of logical conditions in the model, limiting the number of covariates that are used in the model, or limiting the type of models that our algorithms can use in other ways. As long as our models are not so expressive that they can fit random noise, we are in good shape for generalization of our machine learning algorithm.
For instance, we might choose a large value of $K$ for $K$-nearest neighbors, which smooths out our model. We might try to keep our neural network weights close to zero. Perhaps, we might reduce the number of leaves in our decision tree. We must do something to prevent overfitting, which means in practice that we prevent our model class from being overly complex.

\subsection{Regularized loss}
In accordance with the principles of statistical learning theory, a very large number of supervised learning algorithms minimize variations of the following objective, which is a regularized empirical risk. Again, our training data are pairs $\{(x_i,y_i)\}_{i=1}^n$ which are assumed to be drawn i.i.d. from an unknown distribution on $\mathcal{X}\times \mathcal{Y}$.  Here, $x_i$ are real-valued covariates $(\mathcal{X}\subset \R^p)$, $y_i$ are labels. If we are trying to predict the answers to yes/no questions, each $y_i$ could be either +1 or -1. If we are performing regression, the $y_i$s are real-valued. 
The objective is:
\begin{equation}
\min_{f \in F}  \left(\frac{1}{n} \sum_i \textrm{loss}(f(x_i),y_i) + \textrm{complexity}(f)\right).  
\label{basicloss}
\end{equation}
Again, $\textrm{loss}(f(x_i),y_i)$ measures how close our predictions were to the truth (on the training set), with a higher losses giving a penalty for poor predictions. The complexity term encourages the algorithm to choose a low-complexity model. 

Machine learning methods were designed to navigate the tradeoff between overfitting and complexity, and in fact, there are many different types of algorithms that can perform very well on many different types of data. Each of these algorithms uses its own loss function and its own way of measuring complexity, its own class of functions, and its own way of minimizing the loss. In Expression (\ref{basicloss}), the loss function could be the squared loss for regression $(y_i-f(x_i))^2$, or it could be the misclassification error $\mathbf{1}_{y_i \neq \mathrm{sign} (f(x_i))}$ for classification, where we try to create $f$ so that its sign agrees with the true labels $y$. Because the misclassification error is not convex, various algorithms use convex substitutes for this loss function.

\subsection{Algorithms}
\label{subsec:alg}
Let us discuss a few of the most common learning algorithms. 

Classically, \textit{decision trees} have been constructed from the top down \cite{BreimanFrOlSt84,Quinlan93}.  In the example in Figure \ref{fig:decisiontreeadditive}, the algorithm would choose a factor for the top split (e.g., ``age''), based on how useful that factor is for predicting the outcome, then the algorithm would split the data according to that factor.  Then, each part of the split data would be viewed separately, sequentially splitting over and over until there are few observations in each leaf of the tree. However, this approach has started to change with more recent global optimization and mathematical programming algorithms (e.g., see \cite{verwer2019learning,HuRuSe19,angelino2018}), so that all the splits are chosen together so that the overall tree is accurate and sparse. For globally optimal trees, the specific version of (\ref{basicloss}) that we would use is: 
\begin{equation}
\min_{\textrm{tree} \in \textrm{Trees}}  \frac{1}{n} \sum_i \mathbf{1}_{[y_i \neq \textrm{tree}(x_i)]} + C\cdot\textrm{NumberOfLeaves}(\textrm{tree}). 
\label{eq:decision_tree_loss}
\end{equation}
Here, $\textrm{tree}(x_i)$ gives us the prediction from the decision tree. The constant $C$ trades off between accuracy and model size.

Decision trees do not smoothly depend on their parameters, so they are challenging to optimize, even for branch-and-bound methods.
For other kinds of models that depend smoothly on their parameters, the general error term 
$\mathbf{1}_{y_i \neq \mathrm{sign} (f(x_i))}$ is problematic because it is not differentiable with respect to model $f$'s parameters. As a result, many algorithms use differentiable, convex upper bounds instead so that gradient descent methods can be used to reduce the error: if we approximately minimize the upper bound, then hopefully we are also minimizing the classification error. In what follows we will present some of these convex proxies for the loss, for different algorithms.

A common approach in machine learning is to average over many models, creating ensembles of simpler models. The \textit{boosted decision trees} \cite{freund1997decision} algorithm is a prime example of this, which uses weighted sums of decision trees as its function class $\mathcal{F}$.  Boosted decision trees iteratively minimize the convex exponential loss function $\exp(-y_i f(x_i))$ which is an upper bound for the classification error: $$\mathbf{1}_{y_i \neq \mathrm{sign} (f(x_i))}\leq e^{-y_i f(x_i)},$$ 
where $f$ is a weighted sum of decision trees. The exponential loss is a differentiable function of the weights on the trees. This allows the exponential loss to be minimized by coordinate descent. Because the exponential loss is an upper bound for the classification error, minimizing the exponential loss tends to give low values of the classification error.

An additional approach to model ensembling is \textit{random forests}, which average together many different trees rather than reducing a loss function \cite{breiman2001random}. Random forests and boosted decision trees both average over trees, and are very powerful machine learning techniques. 

\textit{Logistic regression} uses the logistic loss, $\log (1+ e^{-y_i f(x_i)})$, which again is an upper bound on the classification error, $$\mathbf{1}_{y_i \neq \mathrm{sign} (f(x_i))}\leq \log (1+ e^{-y_i f(x_i)}),$$ and is equivalent to the negative log-likelihood from statistics. For logistic regression, the set of functions $\mathcal{F}$ is linear models, that is, weighted sums of the covariates, so a function $f$ from set  $\mathcal{F}$ would have this formula: $f(x_i)=\sum_j b_j x_{ij}$. 

\textit{Additive models} are similar to linear models, except that they use weighted non-linear functions of the original variables: $f(x)=\sum_j b_j g_j(x_{\cdot j})$, where $g_j$ can be a non-linear function of variable $j$. Here we denoted $x_{\cdot j}$ as the $j^{\textrm{th}}$ component of point $x$, corresponding to variable $j$.

\begin{figure}
    \centering
    \includegraphics[width=\textwidth]{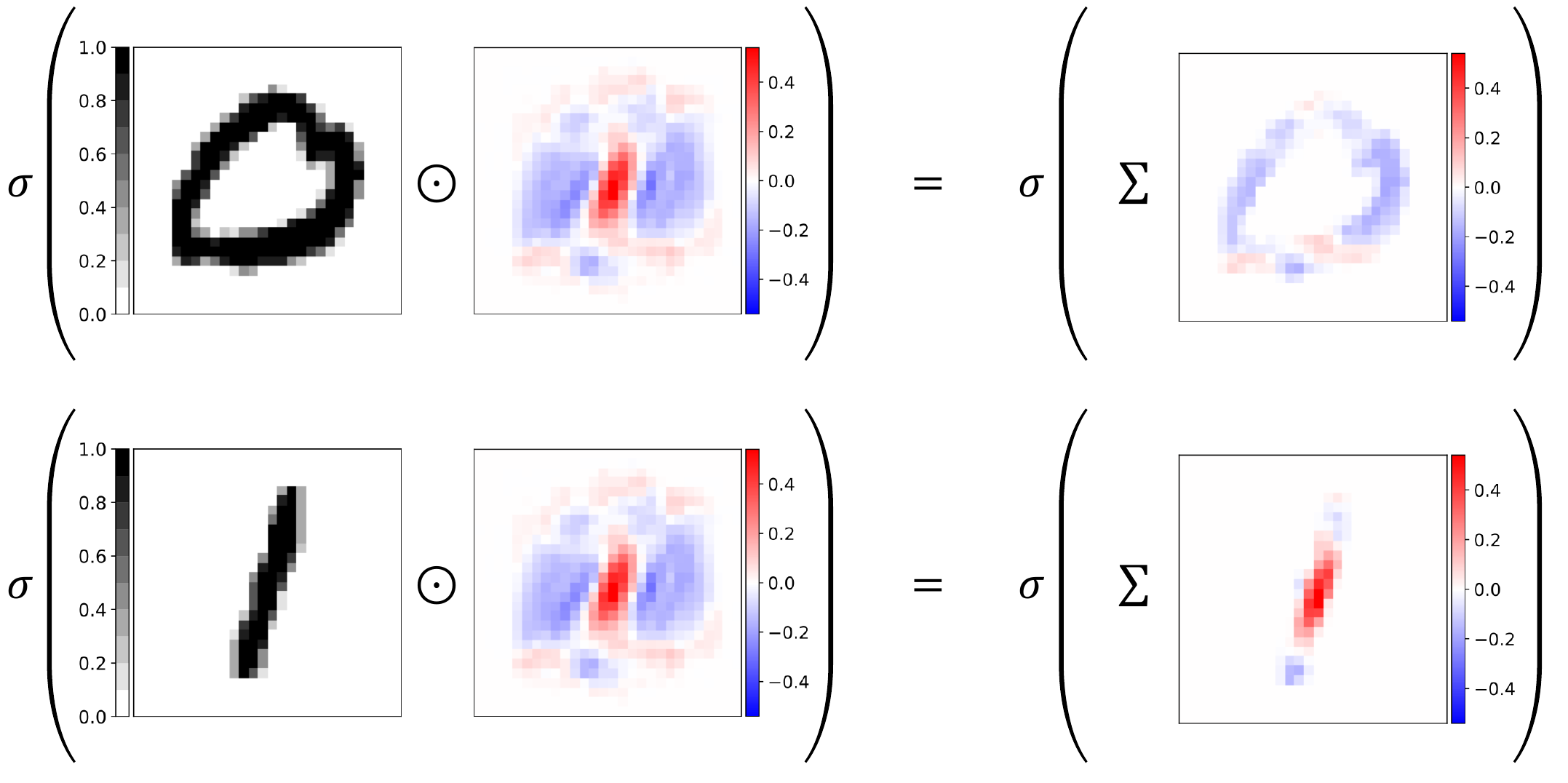}
    \caption{A visualization of logistic regression applied to MNIST data that learns weights to distinguish between the number zero and the number one.  In each digit image, each pixel is represented by a value between 0 and 1, which is then multiplied by the learned weights $\{b_j\}_j$ with the form $f(x_i)=\sum_j b_j x_{ij}$. The values of the $b_j$s are represented in the red-to-blue scale. This data formulation means that each pixel in the digit image is multiplied by the corresponding pixel from the learned weights, yielding the formulation on the right.  To get the final functional form in logistic regression, the pixels are simply summed together.  Note that in these visual tasks, the results will be highly sensitive to slight shifts in the original image, which motivates more complex models such as the \textit{convolutional neural network}. Figure best viewed in color.}
    \label{fig:digits_logistic}
\end{figure}

Figure \ref{fig:digits_logistic} provides an example of logistic regression. Here, each pixel in an image is a variable. The goal is to distinguish handwritten zeros from handwritten ones. The $b_j$s for each pixel $j$ are shown in the red, white, and blue heatmaps. As you can see, when we elementwise multiply an image of a zero with the $b_j$s, the value tends to be negative because of the negative (blue) values around the left and right sides. When we elementwise multiply an image of a one with the $b_j$s, the value tends to be positive because of the positive (red) $b_j$ values in the middle. This way, the sum of $b_jx_{\cdot j}$ is negative for zeros and positive for ones, as shown in Figure  \ref{fig:digits_logistic}.

\textit{Support vector machines} (SVMs) use the hinge loss, $\max(0,1-y_i f(x_i))$, which again is an upper bound on the classification error \cite{vapnik1998statistical}. If the SVM uses a radial basis function kernel (this is the most popular way to use support vector machines), the function class $\mathcal{F}$ is a weighted sum of gaussian functions, each one centered at a different data point, $f(x)=\sum_i b_i e^{-\frac{\|x-x_i\|^2}{\sigma^2}}$, where $\|x-x_i\|$ is the distance between $x$ and training point $x_i$, and $\sigma$ is a parameter of the algorithm. If we set $\sigma$ too small, then each of the terms $e^{-\frac{\|x-x_i\|^2}{\sigma^2}}$ will be close to zero unless $x$ is very close to one of the training points. This means the algorithm cannot generalize beyond the training set, and thus is overfitting the training data. If we set $\sigma$ too large, the $e^{-\frac{\|x-x_i\|^2}{\sigma^2}}$ terms will all be close to 1 regardless of $x$, and the model thus underfits. Figure \ref{fig:alg_comp} shows how a sum of Gaussian functions centered on the data can form a model that fits the data. Here, a  positively weighted Gaussian near a negatively weighted Gaussian creates wiggles in the fitted model.

\textit{Neural networks} also minimize a form of the regularized risk in (\ref{basicloss}). The loss used is usually either the squared loss or the cross-entropy loss.  For a yes/no classification problem, the cross-entropy loss is written as $-y_i\log \sigma (f(x_i))-(1-y_i)\log (1-\sigma(f(x_i)))$, where the sigmoid function is $\sigma(x)=\exp(x)/(1+\exp(x))$. Note that while the cross-entropy loss is typically written in this form, it is mathematically identical to the logistic loss above, and likewise is an upper bound on classification error.  These losses are differentiable and are used in conjunction with gradient methods to learn the model parameters. The function class $\mathcal{F}$ for a neural network is defined by a specific neural architecture. Neural networks use non-linear transformations of weighted sums of variables, and those types of computations are composed together several times.  The first used form of the neural network was the Multi-Layer Perceptron (MLP).  Succinctly, an MLP with a single hidden layer would first define a transformation to the hidden layer, $h_1(x)=g(W_1x+b_1)$, where $g(\cdot)$ is a non-linear function that is applied elementwise, and $W_1$ and $b_1$ are a matrix and a vector, respectively, whose first dimension determines the number of hidden nodes and is a tuning parameter of the network.  The final function form is linear (the same as in logistic regression), where $f(x_i)=\sum_j c_{j} h_1(x_{i})_j$, where $j$ is now the index of the hidden node rather than the $j^\text{th}$ component of point $x$.  More complex neural networks can be generated by either making the network ``wider" (increasing the dimensionality of the hidden nodes) or by having many layers of transformation of the same form (e.g., $f(x_i)=\sum_j c_{j} h_M(\dots h_2(h_1(x_{i})))_j$) which would then have $M$ hidden layers.  The original data have been transformed to be represented by the hidden layers. From this viewpoint, the neural network can be viewed as as extracting features or learning a representation that is useful for classification or regression with a (more traditional) linear model, which is why neural networks are frequently referred to as ``representation learning."  There are several commonly used non-linear functions, such as the previously defined sigmoid function or the Rectified Linear Unit (ReLU), $g(x)=\max(0,x)$. The hyperbolic tangent (Tanh) is also frequently used, which is $g(x)=2\sigma(2x)-1$. The type of layer introduced here is known as ``dense'' or ``fully-connected'' in the neural network literature, and this mathematical form will change in other types of neural networks.

``Deep learning" with neural networks is currently attracting the most attention in the popular media, notably for its inclusion in state-of-the-art artificial intelligence systems for computer vision, speech, language, and board games.  Instead of being based on the previously introduced MLP structure, these impressive systems use layers that explicitly take advantage of repeated structure in data to decrease the complexity of the model.  The two most prevalent examples of this are the \textit{Convolutional Neural Network} (CNN) in image processing and the \textit{Recurrent Neural Network} (RNN) for series data such as audio or natural language. 

The CNN follows the same iterative summarization strategy as the MLP, but each network has distinct mathematical forms for the layers.  The CNN, when applied to images, uses 2-dimensional convolutions to apply the same mathematical operation to a sliding window over the image; this mathematical framework allows a single feature to capture repeated structure over the image, for example horizontal or vertical edges.  Succinctly, this involves specifying a small spatial filter with a different representation for each channel (i.e., ``red,'' ``green,'' or ``blue'' in a color image or human vision), and convolving this filter over each channel in an image (i.e., apply to an image by moving a sliding window), then summing the results over each channel. The output of a convolutional layer will pass through non-linearities or go through down-sampling via ``pooling'' operations, which is then used to form a new ``image'' by stacking the outputs from each filter on one another (applying 10 convolutional filters to a 3-channel color image would result in a 10-channel ``image'' in the hidden layer).
The major benefit of this approach is that when these convolution operations are combined with pooling operations, a network will have built some degree of spatial invariance.  In image analysis, spatial invariance is critical; this can be understood by considering the logistic regression approach in Figure \ref{fig:digits_logistic}, where a small shift in the input image would render the classification system unusable. Instead, we would like to have an approach that gives the same result under shifts of this image; if we move the image a few pixels to the right or left, it should give the same answer.  By using a CNN, this type of spatial invariance can be achieved because a single feature in each layer searches all over the image. The current famous examples in image processing and classification all use various types of CNNs. This has a long history in the analysis of written digits (the celebrated MNIST dataset) \cite{lecun1995learning} and in the ImageNet challenge, which is a large scale classification challenge (one thousand image categories, one million training images). A structure with several convolutional layers followed by the dense layers used in MLPs was what initially led to a breakthrough in the ImageNet challenge \cite{krizhevsky2012imagenet}.

Likewise, the RNN uses layer structures to help incorporate information from the past.  The type of layer used in an RNN takes input both from the features or observations at the current time point and also from the hidden layer of the previous time point.  Using this structure, one can build time- or series-dependencies, critical in applications such as audio processing or natural language processing.  Both of these networks highlight that when layer are custom-designed to utilize specific data properties, they can significantly improve performance.

\subsection{Parameter Tuning}
When using a learning algorithm, the algorithmic parameters must be set. These parameters might include regularization parameters such as the constant $C$ in Equation (\ref{eq:decision_tree_loss}), parameters for how deeply the trees should be grown in boosting and random forests, the parameters of the kernel width $\sigma^2$ for support vector machines, and any other parameters that one might use to tune the algorithm. 

The algorithm's parameters can drastically impact performance (see Figure \ref{fig:alg_comp}).
Almost every learning algorithm has at least one parameter to tune, and sometimes there are many parameters. Tuning of these parameter values can often be an easier path to success than choosing a different machine learning algorithm. The power of tuning parameters should not be underestimated.

A standard technique to tune parameters is cross-validation, where different subsets of the data are used to estimate the predictive power of an algorithm with each possible setting of the parameters. From there, the parameters are selected to maximize expected predictive performance. However,  optimizing even 5 or 6 parameters through cross-validation is extremely difficult, because it essentially involves a grid search for the best parameter values over a 5- or 6-dimensional space.   As we will discuss later, neural networks can have arbitrarily complicated architectures and optimization procedures and are notoriously difficult to tune. Even random forests can be difficult to tune. Boosted decision trees seem to be the most reliable in their ability to generate good models for many different types of data without much tuning.

\section{All machine learning methods perform similarly (with some caveats)}\label{sec:equally}
Neural networks have received a huge amount of attention because of their record-breaking performance on several benchmark computer vision and natural language processing tasks. Indeed, neural networks seem to perform significantly better on many computer vision tasks than other machine learning methods. Computer vision benchmark tasks, however, do not represent the full spectrum of problems that machine learning is used for. 

Surprisingly, there appears to be no practical significant difference in performance between machine learning methods for a \textit{huge} number of data science problems \cite{Razavian15, CaruanaEtAl15, ZengUsRu2017, angelino2018}. In particular, most machine learning methods tend to perform similarly, \emph{if tuned properly}, when the covariates have an inherent meaning as predictor variables (e.g., age, gender, blood pressure) rather than raw measurement values (e.g., raw pixel values from images, raw time points from sound files). 

Computer vision problems are different than most types of problems: natural images live on a narrow manifold in pixel space, whereas most structured data do not. Also neighboring pixels in natural images tend to be closely related to each other. By changing even a few pixels, one can change a natural image to be unnatural (an image that would never occur in nature). Other types of problems (e.g., criminal justice problems) are not nearly as sensitive to small changes in input features. That is, natural images are complicated, but they are complicated in ways that are common among images (e.g., edges, textures). Using these commonalities between images, it is possible to build more useful complex features from the original pixels. Specifically, CNNs use convolutional layers to scan for structure in groups of pixels all over an image, allowing the neural network to utilize this inherent structure. 
In other domains, it is not clear that such additional features would necessarily be helpful. This is possibly why CNNs are so helpful for computer vision but have not been particularly helpful for most other types of data science problems.


Let us consider the vast set of problems for which neural networks do not have an advantage over other methods. For those problems, comparisons between algorithms becomes difficult because there are such small differences between many different algorithms. Each algorithm has several tuning parameters. The differences in tuning techniques can overwhelm the small differences between choice of algorithm. Also, machine learning algorithms are often trained on different distributions than they are tested on, which means small differences in training performance do not translate to exactly the same differences in test performance. This is why ``bake-off'' comparisons between many algorithms on the same dataset might not be particularly helpful, as small differences between algorithms can be meaningless in practice. Training performance can be the tip of the iceberg of what goes into making a machine learning model practical. Issues such as interpretability, importance of variables, fairness, robustness, and so on can be much more important than small differences between algorithms. 

Interestingly, adding more data, adding domain knowledge, or improving the quality of data, can often be much more valuable than switching algorithms or changing tuning procedures; the surprisingly beneficial effect of adding more data has been called ``the unreasonable effectiveness of data'' \cite{halevy2009unreasonable}. 
Adding in domain knowledge in various ways can sometimes be more powerful than anything else: domain knowledge can simplify models tremendously, substitute for lack of data, and help the training procedure towards better solutions.

Thus, the recommendation is simple: if you have classification or regression data with inherently meaningful (non-raw) covariates, then try several different algorithms. If several of them all perform similarly after parameter tuning, use the simplest or most meaningful model. Analyze the model, and try to embed domain knowledge into the next iteration of the model. On the other hand, if there are large performance differences between algorithms, or if your data are raw (e.g., pixels from images or raw time series measurements), you may want to use a neural network. You may not need to train the network yourself, you might be able to download one that has already been trained for a problem similar to the one you are working on.

Lately, a selling point for companies is that they are ``doing deep learning." As we will explain below, neural networks would almost never be the first approach to try on a new dataset, as their training is finicky and difficult, and they tend to be black boxes that are difficult to understand and troubleshoot. We would trust a company more if they claimed to be doing \textit{machine learning}, starting with models that are less likely to overfit and easier to troubleshoot and trust than neural networks. Hence, this claim of ``doing deep learning'' could be a telltale sign that this company may not be conducting careful statistical analysis. We will now go into the challenges of deep learning.

\section{Neural networks are hard to train and weird stuff happens}

As noted in Section \ref{subsec:alg}, the structure of a neural network is that of iterative summarization by multiple hidden layers with the functional form $f(x_i)=\sum_j c_j h_M(\dots h_2(h_1(x_{i})))_j$ for $M$ hidden layers.  Each one of these hidden layers may be the so-called ``dense" layer $h_m(W_mh_{m-1}(\cdot)+b_m)$, where $h_{m-1}(\cdot)$ is the output of the previous layer, or use structures such as convolutional layers that scan the image looking for specific elements (e.g., edges).  Setting up the optimization goal is straightforward, which follows \eqref{basicloss}.  However, optimizing this goal with respect to the networks' parameters is not straightforward, with many settings to choose and many things that can go wrong.  There are two major issues that need to be addressed in practice, which are the network's architecture specification (what are the layers? and how many are there?) and the learning of the network.  We will first discuss the optimization, which is challenging because the objective function is highly non-convex and lacks smooth gradients with the commonly-used ReLU non-linearity. In practice, deep networks are sensitive to initialization, vanishing gradients, numerical issues, and the performance may vary depending on the chosen optimization approach, which will impact the qualities of the found local minima.  

These complications of neural network optimization are in stark contrast to most of the other algorithms discussed earlier. In fact, the learning algorithms for most approaches besides neural networks are typically well-behaved with excellent high-level packages (e.g., Scikit-learn \cite{pedregosa2011scikit}), rendering optimization as a relative afterthought compared to model specification. In particular, most learning algorithms aside from neural networks either have differentiable convex objective functions and use gradient descent for optimization (boosting, random forests, logistic regression), or they use greedy methods (e.g., classical decision trees), or they may use search techniques rather than optimization (classical nearest neighbors), or they rely on solving optimization problems for which optimization theory and solvers exist (linear programming, quadratic programming, semidefinite programming, mixed-integer programming, etc.).  While certainly other intractable problems exist besides neural network optimization, a large divide from classical algorithms occurs when one is working with deep neural networks where the learning algorithm is largely controlled by the user, and the user's choices can have a large impact on which solution is obtained.  

In these deep networks, almost all learning algorithms are based on stochastic gradient methods \cite{bottou2018optimization}, where the gradient is estimated by a small number of data samples (the ``minibatch" size) and a small step is taken to go in the direction opposite to the gradient to decrease the objective. Among the two most common currently-used algorithms are stochastic gradient descent with momentum (SGD-M), which has been used in practice on many top performing neural networks, and the adaptive moments (ADAM) algorithm \cite{kingma2014adam}.  Like all stochastic gradient methods, the SGD-M algorithm first estimates the gradient of the neural network on a small subset of data points.  This gradient estimate is then smoothed by adding to a exponentially decaying sequence of historical gradients before taking a step. Within this general framework, there is significant variability in the individual learning algorithms, minibatch sizes, learning rates (step sizes), when to reduce step sizes, and momentum parameters.



Traditionally, momentum methods have been viewed as techniques to dampen highly oscillatory dimensions, but in the context of deep learning they can be also be viewed as denoising techniques on the gradient estimate.  SGD-M requires the end user to determine the step-size sequence (how and when to decay to smaller step sizes) and momentum parameter.  In the context of well-formed convex problems, there is significant theory underlying how to set these parameters, but since neural networks are non-convex and typically non-smooth objects, this theory is merely a suggestion and is frequently ignored in practice.  One idea is to adapt the step sizes from the historical gradient estimates themselves, leading to many adaptive procedures such as ADAM.  A major potential boon of the ADAM algorithm is that it can be robust to default algorithm settings because it can automatically adapt to curvature and gradient uncertainty to set the step size; however, which optimization algorithm is faster and more robust is highly problem dependent and is under continued debate \cite{wilson2017marginal}, including arguments that each algorithm acquires solutions with different properties, and may generalize differently.  This contrasts sharply with the convex settings of many traditional methods; in deep networks, \textit{how we chooose to optimize impacts what type of solution and generalization performance we achieve}.

Modern neural networks can have millions of parameters arranged in many layers of non-linear transformations, and they can completely separate reasonable sized classification training sets with perfect classification in even small networks. This of course leads to the possibility of overfitting. One common approach towards addressing this is early stopping \cite{bottou2018optimization}, where the optimization algorithm is stopped when the performance estimated on a validation dataset stops improving, rather than when the training loss stops decreasing.  There are numerous studies demonstrating that the choices of learning algorithm, step size, and batch size all impact the generalization of the network, and that stochastic gradient methods are viewed as an implicit regularization when combined with early stopping.  Therefore, in deep learning, the optimization algorithm implicitly acts as a complexity penalty on the parameters, which is currently not well-understood theoretically. This marks a significant departure from most other machine learning algorithms, where the complexity is controlled by tuning parameters, and the optimization terminates based only on the regularized training loss.\footnote{Note that there are some studies on early stopping in boosting as well.}

\begin{figure}
    \centering
    \includegraphics[width=\textwidth]{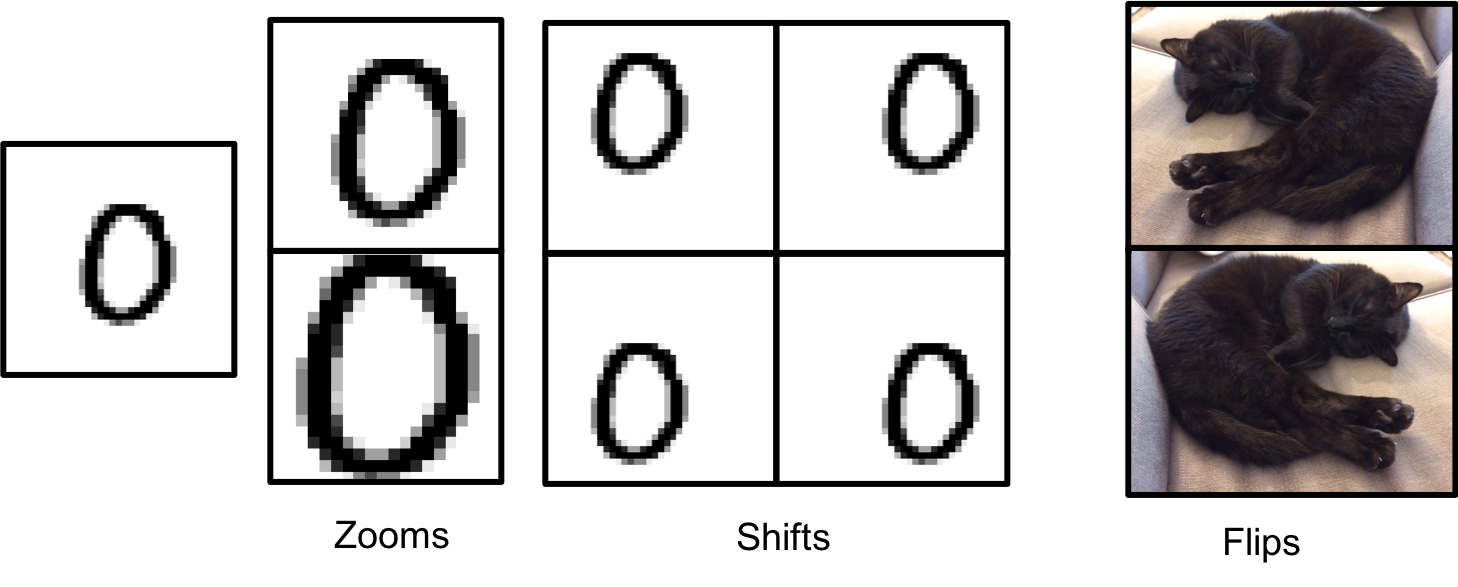}
    \caption{Data augmentation techniques are common in images, which help learned machine learning models be robust to certain types of transformations.  For example, in images, a slight horizontal shift or a slightly larger size should not influence classification predictions.  Likewise, in many natural images flips or reflections should not change the class label of the image. Adding data with these transformations to the training set often leads to improved generalization.  Numerous techniques exist for data augmentation, but often need to be custom-built for new applications, and careful considerations must be made for what augmentation is appropriate (e.g., a reflection on a photo of a cat makes no difference, but reflection does often change the nature of text recognition). }
    \label{fig:data_aug}
\end{figure}

Additionally, neural network performance is dependent on the initialization of the parameters, which is typically generated randomly.  For a new application, one of the most common techniques to boost performance is to initialize a network with a pre-trained model rather than starting from scratch.  This is especially important in the context of medical imaging, where the number of available labeled images for supervised learning is small. To classify medical images, a deep convolutional neural network is first pretrained on natural images, and then is only ``fine-tuned'' on the relevant clinical data, as has been shown in diabetic retinopathy \cite{gulshan2016development} and computer-aided diagnosis in breast cancer \cite{kooi2017large}.  Pre-training in these data examples has been absolutely critical for reaching expert-level performance on these tasks.  However, there is significant variability in what pre-training dataset to use and what parameters to adapt in the fine-tuning, and a theoretical understanding is currently not well developed.  Novel methods to facilitate these types of applications, including transfer learning and domain adaptation \cite{ganin2016domain} are under continued development.  In some situations, there are specific methods that can significantly enhance performance via data augmentation.  For example, in image processing one can include standard image processing tricks such as providing shifted and flipped representations of images, shown in Figure \ref{fig:data_aug}.  Likewise, synthetic data can be incorporated to boost the size of a labeled training dataset \cite{shrivastava2017learning}. This corresponds to a human expert providing additional expertise in the form of additional data.

Overall, deep learning has a set of exciting techniques for modeling data, but there are many open theoretical questions, and non-experts can experience difficulty getting started in deep learning due to the nuance of the learning algorithms alone, let alone specifying a network.  Fortunately, a large community of individuals has started to contribute open-source versions of code that reproduces their results.  This is especially fortunate, as due to the complexity of network definitions and learning algorithms in modern deep networks, it would be extremely difficult to replicate code from reading a paper alone.  However, these standards are not followed by all, as sharing source code is less common in industry; this leads to situations where simply reproducing a result claimed by another organization may be a full project in itself.

\section{If you don't use an interpretable model, you can make bad mistakes.}
\label{sec:interpretable}
The term ``black box models'' refers to predictive models whose calculations are either too complicated for a human to understand or models that are proprietary, whereas interpretable models provide the reasoning process for each prediction. 

Lack of trust is a major issue for implementing machine learning systems in practice. How do we know that a complex black box trained on a static dataset in a particular controlled environment is going to behave as we expect when launched in the wild? Conversely, how do we know our algorithm is not picking up on confounding signals? Recently in a study of deep neural networks on xray images, the networks were found to be looking at some writing indicating the type of xray equipment rather than at the actual medical content of the xray \cite{Zech2018}; this is a bad type of confounding, and similar problems happen in almost all medical datasets and in other types of data as well. Beyond that, how do we know that machine learning models won't create predictions that are so far wrong that they create unsafe conditions for human survival? During the California wildfires of 2018, Google replaced the air quality index of the Environmental Protection Agency with a proprietary black box machine learning model from an outside company. This black box model claimed that pollution levels were safe even when people were reporting a layer of ash on their cars \cite{McGough2018}; this easily could have led to unsafe situations for many people. The questions go on: how do we know whether these models are not biased against a group of people? How do we know whether these models depend on information that is available and trustworthy in practice? How do we know that these models do not inadvertently release private information? These are all complicated issues, but all of them are much easier to deal with if the machine learning model is interpretable to a human expert. In essence, lack of trust is a major issue for machine learning methods, \textit{and it should be}, due to the fact that training data are often \emph{flawed in unknown ways}. Trust in machine learning is not just about trusting the optimization procedure for training the model, it is about trusting the quality of data and how representative the data are of cases where the model would be used in practice.

High stakes decisions should be handled in a very different way than low stakes decisions. The field of machine learning developed essentially for low stakes decisions, such as handwriting recognition, image classification on the internet, and online advertising. In low stakes decisions, mistakes don't cause irreparable harm, and can be made regularly without large repercussions. In contrast, in high stakes decisions, even one false negative or one false positive could be too many. It could lead to a dangerous criminal being granted bail, or a reliable person being denied a loan for a house or new business. For high stakes decisions, black box models should not be used unless absolutely necessary, and it is unclear that it is ever actually necessary to use a black box machine learning model -- for high stakes decisions, \textit{one should always aim to construct an interpretable model that is as accurate as the black box}. 

As we will discuss below, the case of risk scoring models used in the criminal justice system is a case where a complicated proprietary model is used widely (but where there exist alternative simple models that are just as accurate, discussed below). Let us consider the proprietary model COMPAS (Correctional Offender Management Profiling for Alternative Sanctions). COMPAS involves up to 137 factors, which are entered into a computer manually. Manual data entry leads to data quality issues. Because the COMPAS model is proprietary, it is difficult to check that all of the data have been entered correctly. As a result of these typos, there have been some high profile cases where incorrect data entered into the COMPAS survey has led to denial of parole \cite{nyt-computers-crim-justice}. Data quality is also problematic in healthcare applications; if someone mistakenly enters even a single wrong prescription into electronic health records, it can significantly affect predictions.   

Most people using machine learning do not even make an \textit{attempt} to create an interpretable model. There could be many reasons for this, the primary one being that it is harder to construct an interpretable model than simply an accurate one, as discussed below. In other cases, organizations would prefer to keep their models proprietary, as a trade secret, and thus would not want to construct an interpretable model. In other cases, people do not attempt to create interpretable models because they believe that they would need to sacrifice accuracy to gain interpretability. As we discuss below, that probably is not true in most domains. Instead, as we discussed, many people are using black box models for high stakes decisions, which can lead to (and has led to) bad decisions.

\section{Explanations can be misleading and we cannot trust them
}\label{sec:explain}

Lately, there has been substantial interest in trying to explain black boxes. Usually, an ``explanation'' is a separate interpretable model that approximates the predictions of a black box model. The problem with these explanations is that they do not actually explain what the black box is doing, they are instead summary statistics of the predictions. 

Because variables are often correlated, \textit{explanation models often depend on different variables than the original black box model}. This can lead to confusion about what the black box models depend on. For instance, let us consider the important case of ProPublica's analysis of the black box COMPAS model \cite{propublica2016}. ProPublica created a dataset with the ages, criminal histories, gender and race from thousands of individuals, along with COMPAS scores (predictions from the COMPAS black box model used in the justice system). They used a linear model to create an explanation model for COMPAS. In other words, they created a linear approximation of the COMPAS scores on the data to mimic the predictions of COMPAS. Since their model had a race coefficient that was statistically significantly different from zero according to standard regression analysis, they concluded that COMPAS depends on race, even after taking age and criminal history into account. However, their analysis had at least one serious flaw: COMPAS seems to depend non-linearly on age, and once that non-linearity is considered and subtracted out, machine learning algorithms provide no evidence to back ProPublica's claim \cite{RudinWaCo18}. This shows one of the dangers of trying to explain black box models -- the variables that are important for the explanation model (like race) are not always important to the black box.

\begin{figure}
    \centering
    \includegraphics[width=.7\textwidth]{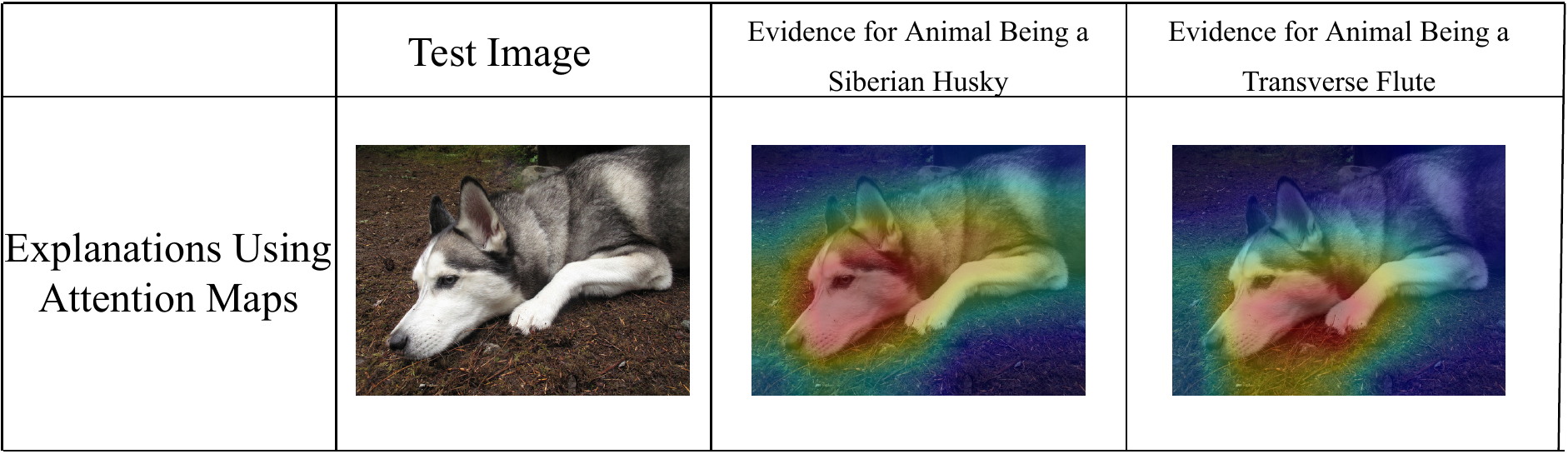}
    \caption{Saliency map. The heatmap in the middle shows the explanation of why this image might be within the Siberian husky class. The heatmap on the right is the explanation of why the image might be classified as a musical instrument. The problem with saliency maps is that their explanations are often difficult to use, and may not provide relevant information. (Figure created by Chaofan Chen. \cite{Rudin19})}
    \label{fig:saliency}
\end{figure}

There are other reasons why one should be cautious about explanations for black box models (see \cite{Rudin19} for more details), namely:
\begin{itemize}
\item \textit{Fidelity}: Explanations cannot fully capture how a decision was made; otherwise the explanation's predictions would be equal to the black box's predictions and one would thus not need the black box at all. If the explanation differs from the black box often enough, one cannot trust the explanation, and thus, one cannot trust the black box.
\item \textit{Double trouble}: The combination of a black box model and an explanation model force the model's developer to troubleshoot two models rather than one. 
\item \textit{Insufficient or confusing explanations}: Sometimes the explanations provide little information about how the black box made its predictions. For instance, for computer vision, attention mechanisms and saliency maps are very popular for showing which pixels the network relies on, but they are seriously flawed as explanations. They often give essentially the same explanations for every possible prediction, as shown in Figure \ref{fig:saliency}. (Also see \cite{Adebayo2018}.)
\end{itemize}

This does not mean that explanations for black boxes are always useless; in fact, sometimes it is clear from the explanation that a neural network classified an object in an image incorrectly because it was looking at a completely different part of the image. However, one does need to be cautious about substituting a black-box-with-explanation for an inherently interpretable model.

\section{It is generally possible to find an accurate-yet-interpretable model, even for neural networks
}\label{sec:equalnn}

Very often, across many applications, it is possible to produce an interpretable model that is approximately as accurate as the best available black box. However, creating an accurate interpretable model can be much more complicated than creating an accurate black box. Consider the possibility of changing our optimization algorithm from (\ref{basicloss}) so that it incorporates a constraint on the model's interpretability. These interpretability penalty constraints could change the problem entirely, making it much more difficult to solve. 
\begin{eqnarray}
   &&\min_f \frac{1}{n}\sum_{i=1}^{n}\textrm{loss}(f(x_i), y_i) \nonumber\\
   &&\text{s.t. InterpretabilityPenalty}(f) < C.\nonumber
\end{eqnarray}

Let us give an example.
We can start with standard logistic regression, and add constraints that the model is sparse and has only integer coefficients. This optimization problem provides a recipe for creating a \textit{scoring system}, which is a sparse linear model with integer coefficients (see \cite{UstunRu2017KDD}):  
\begin{eqnarray}
   &&\min_{\bm{\lambda}} \sum_{i=1}^n \log(1+e^{-y_i \sum_j x_{ij}\lambda_j}) \nonumber\\
   &&\text{s.t.}\forall j,\; \lambda_j\in\{-10,-9,...,0,...,9,10\},\;\;\; \|\bm{\lambda}\|_0 < C.\label{eqn:riskslim}
\end{eqnarray}
Here, $\|\bm\lambda\|_0$ is the number of nonzero elements of $\bm\lambda$.
The interpretability constraints force the domain to be discrete, which means the whole problem becomes a mixed integer non-linear program (MINLP), which is very difficult to solve. 

\begin{figure}[h]
    \centering
    \includegraphics[width=0.7\textwidth]{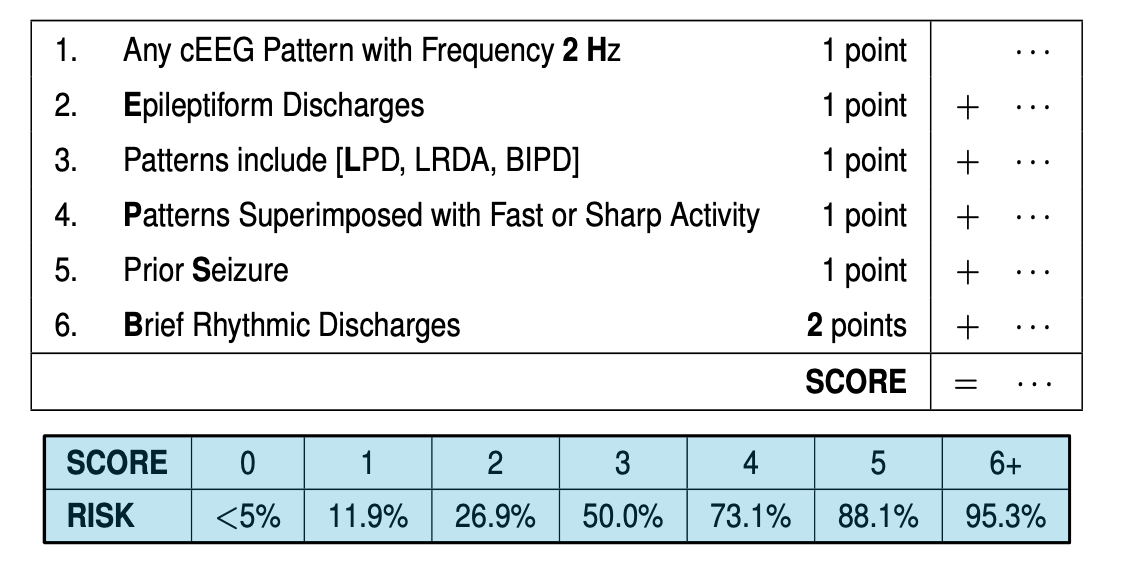}
    \caption{The 2HELPS2B score for predicting seizures in ICU patients \cite{StruckEtAl2017}. The total number of points is translated into risk of near-term seizure using the table along the bottom. The factors within the model were chosen (by an algorithm) from a large collection of possible factors. The point scores were also chosen by the algorithm. The risk factors in the model are known risk factors that neurologists are trained to recognize.
    \label{fig:2HELPS2B}}
\end{figure}

For many problems, these additional constraints do not change the objective value of the optimal solution, and yet they make the models much more interpretable and easy to use. Figure \ref{fig:2HELPS2B} shows a scoring system that is a solution to (\ref{eqn:riskslim}) from Struck et al. \cite{StruckEtAl2017}. The model in Figure \ref{fig:2HELPS2B} was created by the RiskSLIM algorithm (Risk Supersparse Linear Integer Models) which is a specialized technique that combines cutting planes with branch-and-bound methodology to solve (\ref{eqn:riskslim}).


Scoring systems are useful for certain problems, and there are other forms of interpretability for different domains. Interpretability could be challenging to define for some domains, such as computer vision or speech. There have been some efforts to make neural network models interpretable. If the neural network is interpretable, we do not need a posthoc explanation of what it is doing \cite{ChenEtAl18,LiEtAl18}. We will give two examples below. It is important to note that there is not a single definition of interpretability across domains. There is no reason to expect there to be a single performance measure of interpretability, since there is not a single performance measure for prediction performance across domains; accuracy, false positive rate, area under the ROC curve, F-score, discounted cumulative gain, etc., are performance measures used in various domains. There is no reason that two different domains need to use either the same prediction performance criteria or the same interpretability criteria. In the two examples below we consider specific definitions of interpretability for computer vision for natural images (Section \ref{sec:vision}) and neurological electrophysiological data (Section \ref{sec:electro}). 

\subsection{An example of an interpretable neural network for computer vision}\label{sec:vision}

We will discuss a new interpretable neural network approach. The approach is called ``This looks like that" (TLLT) because it classifies a new image based on how similar its parts look like prototypical parts of past images within the training set. 
TLLT's interpretability measure was designed to mimic explanations provided by humans for difficult visual classification tasks. It uses case-based reasoning. Birdwatchers might determine that a bird is a clay-colored sparrow because its head looks like that of a prototypical clay-colored sparrow, its throat and breast resemble that of a (possibly different) prototypical clay-colored sparrow, the pattern on its feathers resembles that of another prototypical clay-colored sparrow, and the evidence of all of these observations is enough to determine the class label of the bird. TLLT is designed to reason about cases in this way as well. The network points out parts of a test image that resemble parts of past cases. It sums this evidence into a single score for each class, and chooses the class with the highest score.

\begin{figure}
    \centering
    \includegraphics[width=0.5\textwidth]{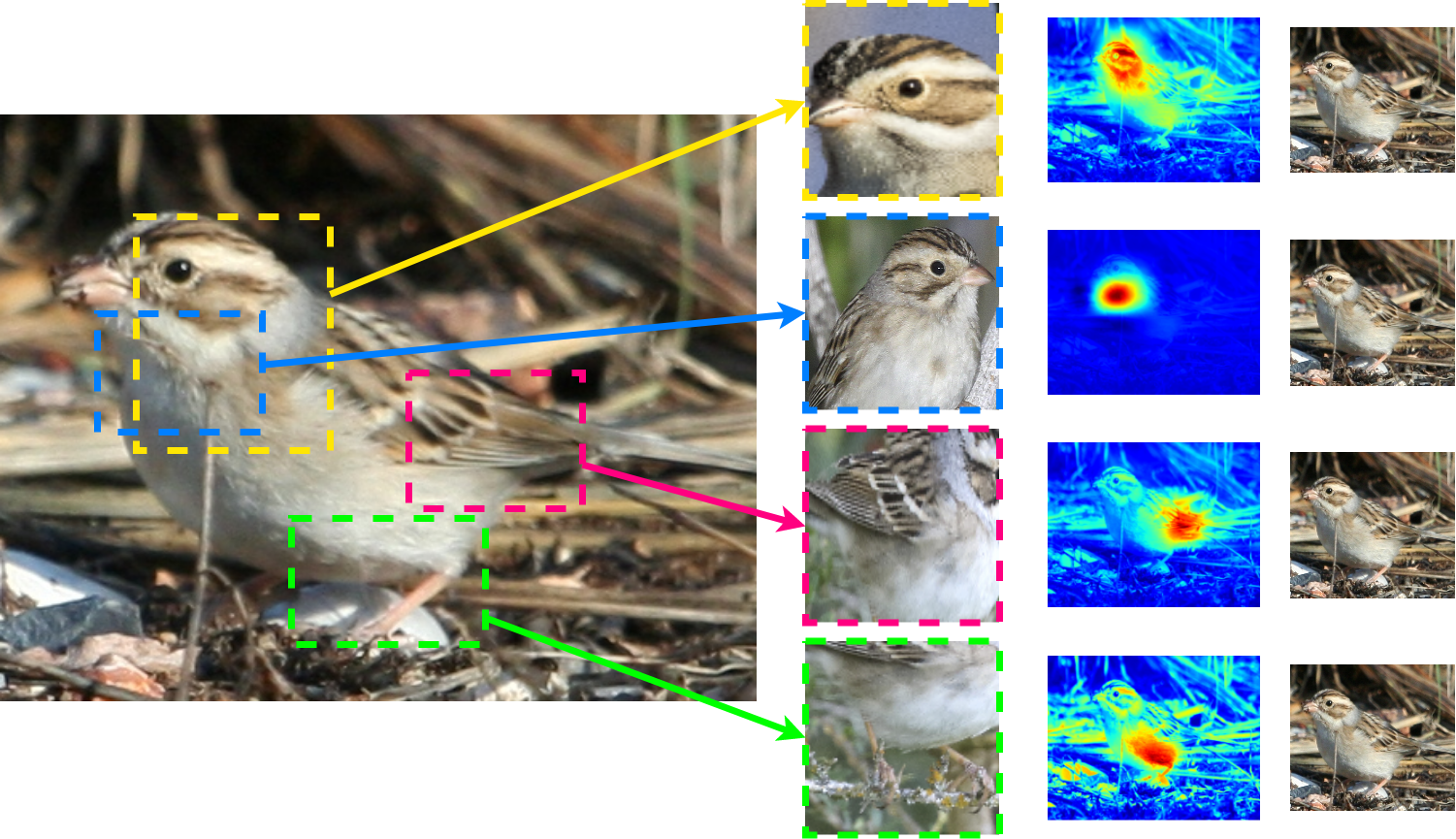}
    \caption{A prediction made by the ``This looks like that" interpretable neural network on a test image \cite{ChenEtAl18}. The network is highlighting certain parts of the image that are useful for determining its predicted class. It is explaining to the user that it classified the image as a clay-colored sparrow because parts of the test image look like prototypical parts of clay-colored sparrows that it has seen before. The right column is just a copy of the original test image. The second-to-right column highlights parts of the test image that look like the prototypical past cases from the training set in the third-to-right column. For instance, the head of the test bird is highlighted in the top right images, which the network thinks is similar to the prototypical clay-colored sparrow's head in the middle at the top. (Image created by Alina Barnett. \cite{Rudin19})}
    \label{fig:TLLT}
\end{figure}

Figure \ref{fig:TLLT} shows an example of a prediction made by the TLLT neural network. The network points out similarities to past prototypical cases, and highlights where comparisons are made between parts of the test image and prototypical parts of training images. The rationale for the prediction provided in Figure \ref{fig:TLLT} is not aiming to approximate (explain) a black box prediction; instead, the network is actually using the evidence it gathers from the prototypical past cases to make its decision about the test image. More details about TLLT are provided in \cite{ChenEtAl18}. The reasoning process shown in Figure \ref{fig:TLLT} contrasts with the saliency map posthoc analysis shown in Figure \ref{fig:saliency}.

TLLT performs a type of cased-based reasoning, similar to the way real-estate agents might price a new house based on comparable houses that have recently been sold, or how doctors might reason about a new patient in terms of similar past cases. If the physicians know which cases the network is comparing the present case to, and how the cases are being compared, the physicians can decide for themselves whether the prediction is able to be trusted.

\subsection{An example of an interpretable neural network model for electrophysiological data}\label{sec:electro}
\begin{figure}
    \centering
    \includegraphics[width=\textwidth]{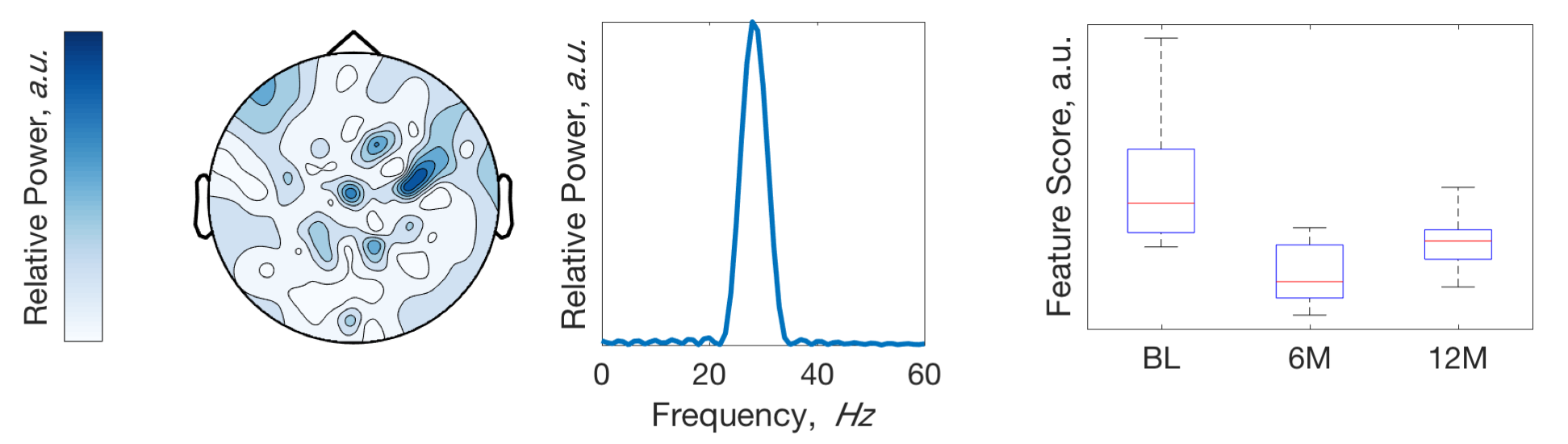}
    \caption{An example of a transparent neural network for electrophysiology data from \cite{li2017targeting}.  A network is trained to identify what stage of data collection recording is from: baseline (BL), 6 months after treatment (6M), and 12 months after treatment (12M), with the goal to learn neural differences that occur after treatment.  The three subfigures above document the properties of one of the ten features learned in the neural network.  (Left) This figure shows relative power in arbitrary units (a.u.) defined by the learned variables in the network.  This visualization is from the popular Fieldtrip \cite{oostenveld2011fieldtrip} package, and is a common way to represent neural signal distribution over the brain.  This figure reveals that the majority of the signal strength used in this feature to make decisions is in the midbrain.  (Middle) This figure shows the frequency range used by the learned filter defined by the learned variables. Frequency bands are a commonly used feature in the field (i.e., brain waves in different frequency bands), but are typically set rather than learned.  (Right) To demonstrate the effect of the learned feature, one can visualize its ``score,'' or value from each data sample, at the three different tested conditions.  These scores are calculated by evaluating how the network operates on different input samples.  This feature has higher values at baseline than it does at 6 or 12 months after treatment.}
    \label{fig:eeg_interpretable}
\end{figure}

One approach to building transparent neural networks in scientific tasks is to create bespoke models that can capture domain-relevant qualities.  Such an approach requires close collaboration between methodological experts and domain experts.  For example, when dealing with electrophysiological data, one approach is to parameterize and constrain the filters in a convolutional neural network so that it utilizes frequency patterns and spatial coherence \cite{li2017targeting}.  These types of properties are commonly used in electroencephalography (EEG) analysis, blending the strengths of neural networks with traditionally used features.  In that approach, the learned features can be visualized in the context of known approaches in the field to allow direct comparison with more traditional ways of defining features.  A critical consideration when communicating results is to put representations into familiar formats, which can be done by visualizing learned variables using one of the most common software packages \cite{oostenveld2011fieldtrip}. An example of this visualization can be seen in Figure \ref{fig:eeg_interpretable}.

One can imbue machine learning models with other properties besides interpretability, as we discuss next.

\section{Special properties such as decision making, fairness, or robustness must be built in}\label{sec:othercontraints}

When we train our algorithms to minimize loss, the approaches work well at optimizing this metric; however, they only mininize this metric.  If the end goal is to understand predictive uncertainty, have a certain type of robustness, or any other special properties, then the loss function does not necessarily match our goals.  Instead, the full set of goals need to be directly built into the algorithms and addressed either during learning, model selection, or in post-hoc corrections.

A large issue when using algorithms in real-world situations is whether the automated decision is ``fair.''  Because historical data used to train algorithms often contain implicit bias, this implicit bias will be learned the the algorithm.  There are many famous cases of this, including how an automated hiring algorithm has gender bias when recommending top resumes for tech jobs \cite{hiringbias}.  One way to address this challenge is to determine an algorithmic definition of fairness, and enforce that condition while learning. Currently, no definition of fairness has been universally agreed upon. There are many ways in which fairness could be defined, for example that an algorithm recommends an equal share of males and females, that the features used for decision making are not predictive of gender \cite{louizos2015variational}, or that the false positive rates are similar for different groups \cite{berk2018fairness}.  Many other definitions exist. Performing a post-hoc correction on any of these conditions is challenging, because the default loss function does not match the fairness goal. Instead, these mathematical definitions of fairness must be enforced while learning the algorithm, a significant departure from traditional learning frameworks. 

In addition, performance metrics do not necessarily match the decision-making goal of an algorithm.  One example of this is accuracy, which is dependent on our decision threshold $\tau$, which states that we only predict an example $i$ is positive if the function $f(x_i)>\tau$.  The decision threshold is typically 0 unless it has been adjusted.  As a hypothetical example of why this decision threshold can matter in practice, consider a cancer screening test with a 99\% true positive rate (99\% of actually positive cases are predicted as positive) with a 1\% false positive rate (99\% of actually negative cases are predicted as negative). Largely, this sounds like a reasonable test that will be 99\% accurate no matter what.  However, if we applied this test to the population at large, where population cancer rates are less than 1\%, simply stating that no one has cancer would be more \textit{accurate} than our test above.  Stating that all examples are negative corresponds to increasing this decision threshold $\tau$ to a much larger value, so that no data examples would pass the threshold.  While generally one would only apply such a screen to at-risk patients where the data would be more balanced, the point is that looking at accuracy alone may convince us that we would prefer our trivial prediction to the above informative test.  For model evaluation in binary cases, the issue that accuracy is dependent on the decision threshold is typically addressed by looking at ranking metrics that do not require a threshold, such as the Area Under the Curve (AUC), weighted AUC, or Concordance-index; this approach is useful in model comparison but does not solve the issue of setting thresholds in practice. Note that this issue is certainly not limited to machine learning, and appears in all areas of statistical modeling.

When making a decision, we are concerned about the utility of the decision instead of accuracy. Determining what the decision threshold should be in practice depends on making a value judgement about the benefits of detecting true cases versus the penalty for claiming a false positive, and then a decision threshold can be chosen to maximize the utility of the problem.  In a clinical setting, this threshold may be chosen by an determining acceptable minimum true positive rate or maximum false positive rate rather than maximizing accuracy.  In binary cases, this is relatively straightforward. However, choosing thresholds on specific outcomes is less straightforward when dealing with multi-class or regression problems.

\begin{figure}
    \centering
    \includegraphics[width=.45\textwidth]{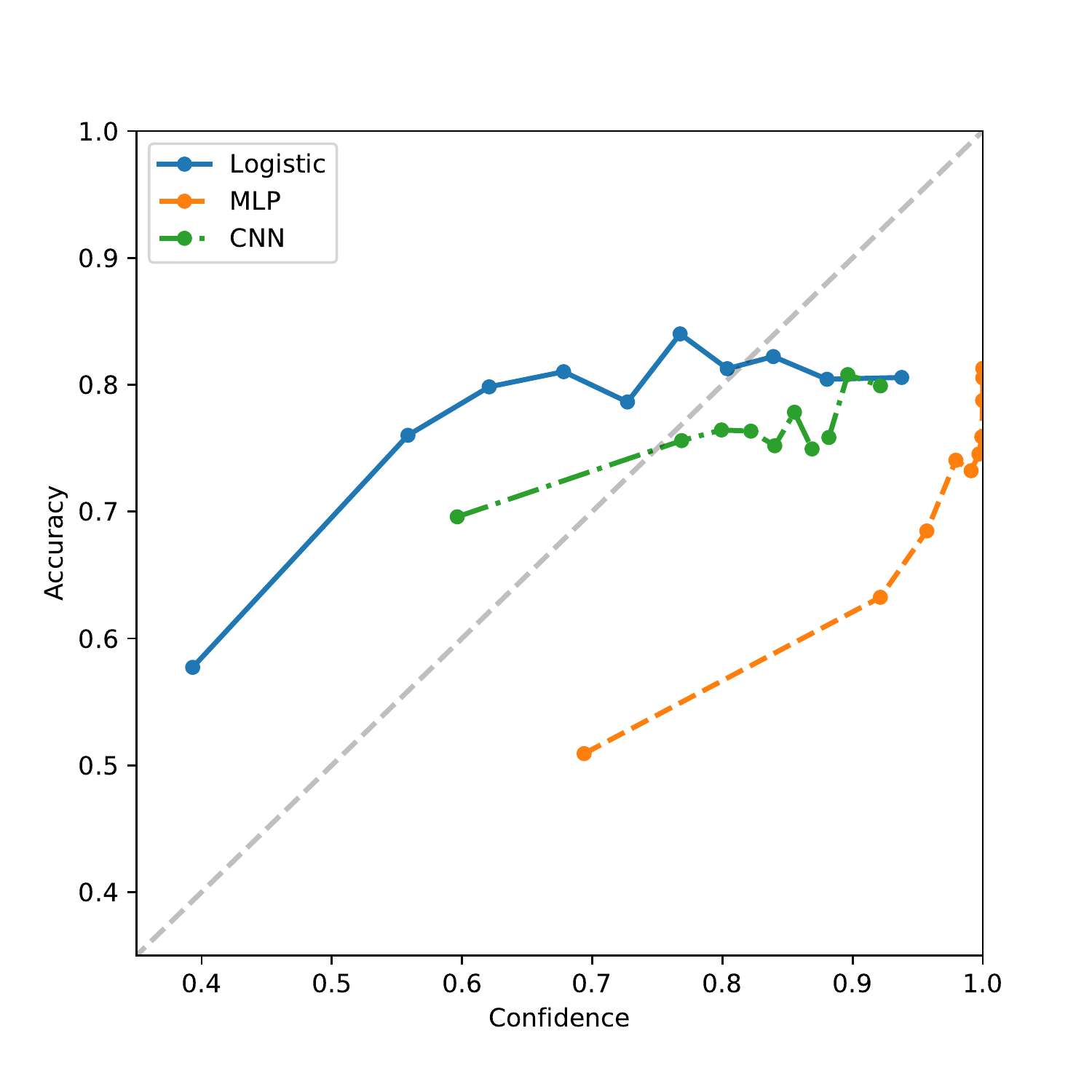}
    \caption{The results of training three different approaches (Logistic Regression, Multi-Layer Perceptron, and a Convolutional Neural Network) on the MNIST dataset where 20\% of the labels are randomly permuted. If the confidence levels on the validation samples were well calibrated, then they should exactly match the inferred accuracy.  However, on these data, no method captures confidence well out-of-the-box, and would require post-hoc adjustments to align the confidence with reality.}
    \label{fig:calibration}
\end{figure}

In the context of decision making, we are often concerned about statistical calibration of results.  Succinctly, if a probabilistic algorithm states that there is a 95\% probability of a positive outcome, is that confidence level trustworthy?  In many complicated machine learning algorithms, and especially neural networks, \emph{it is not} when applied out-of-the-box.  One way to measure confidence intervals is to recalibrate: take all observations with 90\% confidence towards the positive class, and evaluate how many of them actually have positive labels.  This can be repeated for all levels of confidence, which we show in Figure \ref{fig:calibration}.  In that example, it is clear that a deep neural network can become poorly aligned with reality. If calibration is a desired property, then post-hoc calibration methods exist for realigning decision confidence with additional hold-out data \cite{platt1999probabilistic}, but this has to be specifically incorporated and appropriately modified for any data processing pipeline.  In regression, it can be even more challenging to calibrate uncertainties, especially in the context of heteroskedastic errors.

A separate issue is that our data samples may not be representative of the real-world population. To adjust the model to be appropriate for a new population, we might choose a weighting scheme or subsample of the data to mimic the correct population \cite{batista2004study}.
It usually is not enough to simply alter the loss function to target alternative goals. Instead, the way the data are fed to the algorithm or the way the algorithm is trained usually needs to be modified. This requires an end-user to understand how to appropriately modify methods to target alternative goals.

In addition to decision making, we often care about other properties, such as robustness or interpretability.  We discussed interpretability earlier, but the concept of robustness can be rather vague and requires a mathematical definition of robustness and a modification of the training and methodology.  One approach to robustness is based on the assumption that simpler models should perform better when the newly tested data contain slight distributional shifts from the training data, which can be made rigorous in the context of statistical learning theory \cite{ben2010theory}.  However, determining how to reduce complexity that could potentially improve performance on shifted test data is a difficult problem, and is highly related to problems in domain adaptation and transfer learning.
 
Overall, machine learning methods are now at the point where many software packages are able to train models to optimize standard metrics.  The training procedures and model selection techniques are well-matched to this goal.  However, it is important to remember that these methods are typically only designed to match that goal, and one should consider how well those metrics are matched to the scientific or predictive question being asked of the data.

\section{Causal inference is different than prediction (correlation is not causation)}\label{sec:causal}
Machine learning tools have started to be used for causal inference applications, where the goal is to determine the effect of a treatment $t$ on an outcome $y$, for a unit with covariates $x$. Specifically in the simplest setting of the potential outcomes framework of causal inference, let us consider a medical example, where each unit is a patient having covariate vector $x$ which includes the patients' medical histories and demographic information, treatment $t$ is 1 if the patient took a particular drug and 0 if the patient did not take the drug (patients with $t$=0 are in the control group), and outcome $y$ is 1 if the patient had a stroke the following year, and 0 if there was no stroke. Assume there is no unmeasured confounding (no important unmeasured variables). In that case, we hope to predict the treatment effect $E_{Y}(Y|T=1,X=x) - E_{Y}(Y|T=0,X=x)$ for each possible covariate vector $x$. This quantity is called the conditional average treatment effect (CATE). (Here we use capital letters to represent random variables.)


Both $E_{Y}(Y|T=1,X=x)$ and $E_{Y}(Y|T=0,X=x)$ can be estimated using machine learning, however their difference cannot be used as an estimate of a treatment effect without several important assumptions. The assumptions vary from problem to problem, but a typical set of assumptions is:
\textit{strong ignorability} (that we have not left out any important covariates, and that the probability of being treated is between 0 and 1 but not either exactly 0 or 1), and \textit{SUTVA} (the stable unit treatment value assumption, that the outcome of one unit does not affect the treatment of another). One of the current methods to compute treatment effects is using the machine learning method BART (Bayesian Additive Regression Trees) due to its high quality performance \cite{Hill2017,Wendling2017}. 

BART models are not interpretable. For interpretability, \textit{matching methods} are a good choice, where treatment units are matched to similar control units. Matching methods are a form of case-based reasoning. Ideally, treatment units would be matched on all covariates to control units, as if we were conducting a randomized controlled trial on identical twins. However, in reality, one cannot usually find a match on all covariates. Some modern matching methods aim to find matches on as many relevant covariates as possible using machine learning (e.g., \cite{DiengEtAl2019}). Once units are assigned to matched groups, an estimated treatment effect would be the difference in outcomes between treated and control units in each group: $\hat{y}^{\textrm{treated}}-\hat{y}^{\textrm{control}}$, where $\hat{y}^{\textrm{treated}}$ is the average outcome for treated units in the matched group, and $\hat{y}^{\textrm{control}}$ is the average outcome for control units in the matched group. After these estimates of treatment effect are calculated, they can be smoothed using any machine learning method (e.g., linear regression).


There are several common mistakes made in causal inference analysis. One common mistake involves \textit{leakage}, which results from problems with correlation versus causation. Leakage in data occurs when information from the future (the information we are trying to predict, namely the label $y$) is accidentally encoded within the features (the predictor variables $x$). Consider the example of predicting sepsis in hospitalized patients. We would predict a sepsis diagnosis at a future time (say 24 hours after time $t$), given data from before time $t$, using machine learning. Let us say that doctors treat patients for sepsis before the official timestamp of their diagnosis. In that case, many patients in our dataset will have sepsis treatment drugs recorded before being diagnosed. In that case, a machine learning algorithm could easily predict a future diagnosis of sepsis: patients who are given sepsis treatment drugs are highly likely to be diagnosed with sepsis within 24 hours. While this example may seem obvious and farfetched, it (unfortunately) is not. 

The existence of unmeasured confounders would violate the  ignorability assumption for causal inference. Let us give a related example for healthcare by Caruana et al. \cite{CaruanaEtAl15}, who note that having asthma seems to lower the risk of dying from pneumonia. They point out that this is a reflection of the fact that asthma patients and non-asthma patients receive different levels of care; asthma patients are admitted to the Intensive Care Unit more often, and receive a higher level of care, reducing their risk of death. Thus, in order to perform effective causal inference, it would be critical to record the level of care given to each patient in order to see that asthma increases risk of death from pneumonia.

Variable importance analysis cannot be used for causal inference without strong assumptions. A typical way to compute the importance of a variable $v$ is \textit{permutation importance}, where we compute the empirical risk EmpRisk twice: once with our data as usual, and once with a perturbed dataset where the variable $v$ has been replaced with noise. If the values of EmpRisk differ between the clean and noisy versions of the data, it indicates that variable $v$ is important to the model. Researchers often claim that important variables are also causal. Unfortunately this is not true in general. Let us go back to the example of recidivism for criminal justice, as discussed earlier. Since age and race are correlated in all of our criminal justice datasets, one can construct two equally accurate models: one that heavily depends on age and criminal history (and not at all on race), or one that depends heavily on race, but not as much on age. A na\"ive researcher might see only the second model and claim that race is causal for recidivism, despite the fact that neither one of these models would imply that race is a causal factor for recidivism \cite{FisherRuDo18}.


\section{There is a method to the madness of deep neural network architectures, but not always}

An outsider reading recent deep learning papers might believe that there are seemingly infinitely many different network formulations used with no basis to choose among them.  However, over the past decade many specific features of networks have become increasingly common, so in reality the vast majority of networks are using highly related architectures. One of the primary focuses has been on designing improved ``building blocks,'' which are then used to construct a complete architecture by linking these building blocks together. In this way, the simple layers used in classic neural networks are being replaced by layers of these building blocks, each dependent on the goals of the network.  While definitive theory is still elusive in deep learning, the community is building a better understanding of when and why different network structures work well based on theory, heuristics, and a large amount of empirical evidence.  Furthermore, many different applications are now using structures such that it is easy to swap building blocks or even an entire type of network.

CNNs, in particular, have shown incredible advances in image analysis on benchmark challenges \cite{krizhevsky2012imagenet}.
While initial CNN results were extraordinarily promising in image analysis, there have been several recent developments in the network architecture that have dramatically enhanced performance. One key idea repeatedly exploited in recent literature is that of ``residual'' connections \cite{he2016deep}. To briefly explain the idea, first note that stacking layer after layer to form a very deep neural network creates an object that is extremely difficult to optimize, both because of the extraordinarily large set of parameters to learn and because gradients must be propagated across all the layers to construct a learning signal.  These challenges limited convolutional networks for image analysis to roughly $\sim$30 layers, with most networks using significantly fewer layers.  An approach to address this challenge is to create a hybrid structure with both a deep network structure capable of capturing complex behavior and shorter input-output paths to capture the broader activity and facilitate learning.  One way to do this to modify the building blocks of a Convolutional Neural Network to have functions like $h(x)=x+f(x)$, where $f(x)$ may have multiple convolutional layers and $x$ is the input data.  By using this type of building block, we can maintain a much shorter path between the input and output of our network while allowing the deeper structure to provide increased capacity.   This type of residual network approach has facilitated much deeper networks than previously possible, including up to 1000 layers, and the Resnet-50 (50-layers) and Resnet-101 versions of this network are currently extremely popular.  Residual connections are not the only way for forward development, and there has been significant efforts to build other properties into the network, including the popular MobileNet approach to use energy-efficient designs \cite{howard2017mobilenets}.

There is a significant question as to why so many research papers use specific and repeated architectures. As discussed previously, training convolutional networks from scratch is both computationally and methodologically challenging, especially in the context of limited data.  Therefore, instead of training a network from scratch, one uses ``pre-trained'' networks, which usually refers to weights that have been learned carefully on the ImageNet dataset, and then using those weights as an initialization for a new customized network.  This ``transfer learning'' strategy has shown incredible promise in a wide variety of applications, including situations that seem quite distinct from natural images, such as in medical imaging \cite{gulshan2016development}.  Therefore, an end-user is, in practice, limited to networks with available pre-trained weights when working on applications with comparatively little data. This reality has resulted in standardizing and condensing the set of popular architectures.  Deep learning coding frameworks are all designed to facilitate the incorporation of pre-trained weights and these standard network designs.

Depending on the type of data, we may want to use different types of networks and building blocks.  When we consider the analysis of series data, such as a time-series or a series of words (natural language processing), we generally use a Recurrent Neural Network (RNN), although CNNs are also used in these tasks.  The fundamental idea of an RNN is that the architecture contains both layers to build depth at each entry in the sequence, but also connections to the past.  In fact, due to the recurring connection to the past, a ``simple'' RNN with only a single learned layer can be viewed as an infinitely deep network.  Similar to our discussions on CNNs, a key idea is to base an RNN architecture on different building blocks, with the two most common building blocks being the Long-Short Term Memory (LSTM) \cite{hochreiter1997long} and the Gated Recurrent Unit (GRU), each of which contains several connections and non-linearities.  Most of the development in RNN architectures is in how to combine these building blocks into useful structures for the task at hand.  One of the most famous approaches is that of the sequence-to-sequence model \cite{sutskever2014sequence}, which is useful for applications such as text translation.  In such a model, an ``encoder" RNN consists of stacked building blocks, such as the LSTM, that connect to a fixed-length vector.  This fixed-length vector can then be used to initialize the stacked building blocks of a ``decoder" RNN that is designed to predict or generate a sequence of text.  One of the reasons that such a model is so popular in the literature is because of its interoperability with other networks.  Instead of looking at text translation, an image-to-text generation procedure can be developed by swapping the encoder RNN with a CNN applied to images \cite{pu2016variational}.  Once we have component networks and building blocks, we can facilitate new applications by swapping and combining parts of networks.

Finally, the way we structure our inputs to deep networks matters.  Instead of presenting raw data formats, alternative representations can be very useful.  A prime example of this idea is the concept of \textit{word embeddings} \cite{mikolov2013distributed}.  
Word embeddings represent each word in a language by a vector in a high-dimensional space (typically between 300 and 400 dimensions). This encoding allows us to capture relationships between words. 
Words that are highly similar will have similar vectors (``good'' and ``great'' are highly related words, and would be close in embedded space), with semantic relationships ideally maintained, meaning that adding the difference in the word vectors between ``man'' and ``woman'' to ``king'' will match the word vector for ``queen.''  Remarkably these relationships sometimes hold in practice for learned word embeddings.  Replacing natural text with the word embeddings before sending the text as input to an RNN can drastically reduce the amount that needs to be learned in the network, greatly enhancing performance.

Overall, neural network architectures may seem to be a smorgasbord of possibilities, but network structures are becoming more standard with the ability to relatively easily swap out network components for alternative network components or building blocks.  The use of pre-trained networks and word embeddings can drastically enhance performance, especially in the presence of limited data.  In all of these networks, there is a trade-off with complexity.  The simpler the network, the easier to learn with fewer necessary computational and energy resources; however, a well-learned complex network on a large data set can exchange impressive performance gains for additional incurred computational cost \cite{huang2017speed}.

\section{It is a myth that machine learning can do anything}

In many specialized and highly publicized cases, current machine learning methods are outperforming humans.  A prime example of this is the famous ImageNet challenge for image recognition, where human-level performance was achieved in 2015 \cite{he2015delving}, and the estimated error rates are half of 2015 levels as of writing.  However, this is far from the only example, with the highly publicized AlphaGo \cite{silver2016mastering} using CNNs as a major component of their learning framework to help beat the world champion Lee Sedol in the game of Go 4-1 in 2016.  Incorporating machine learning technology into text translation (``neural machine translation'') significantly reduced translation errors in Google Translate ($>60\%$ reduction) \cite{wu2016google}.  In recent years, we have seen incredible progress in machine learning in many additional domains, with novel state-of-the-art systems in speech recognition, object detection, question answering, etc., and deep learning is one of the most promising techniques in artificial intelligence.  This leads to a simple question: what are the major obstacles in building machine learning systems for anything?

While machine learning is promising, it can't solve everything. An algorithm that achieves human-level performance on image recognition in the ImageNet challenge can identify categories labeled in the ImageNet dataset, but cannot necessarily do anything beyond that. Current systems are incapable of reasoning the way humans do about language and images. As a result, for these domains,
machine learning, and especially deep learning, is data hungry.  The famed ImageNet database started with one million images, and the dataset is routinely augmented by additional labeled examples to train more and more complicated networks.  In natural language processing, current networks are pretrained on a corpus of 3.3 \textit{billion} words \cite{devlin2018bert}.  In contrast, in an early stage clinical trial, we rarely even have a hundred outcomes.
Data likewise are rare in many other scientific domains.  Blindly applying large-scale deep networks or other complex machine learning techniques to these applications does not generally help us predict or understand the data better over classical techniques.

 The gap between training data and data in the real world can cause significant problems for machine learning methods, including time-evolving systems, as we discussed above. This problem is a fundamental issue in machine learning and statistics--how much do our data represent the real world? Performance on the famous machine learning benchmarks, while incredibly impressive, is largely evaluated on testing data that is nearly identical in distribution to the training data. This gap between training and test data can be exploited to cause terrible problems for AI systems, as shown by recent adversarial attacks on commonly used deep networks. Researchers have found that computer vision systems are brittle: modifying a single pixel can completely alter an algorithm's understanding of an image \cite{su2019one}, and adding a small sticker on a stop sign or at an intersection can fool even a modern industrial computer vision system for self-driving vehicles. In addition, policies and data change over time.  If we train on last year's data, will the machine learning model be applicable to the data collected today?  This is a challenging question to answer in practice.

Deep learning is also starting to show a remarkable ability to generate data.  This type of approach, known as a generative model (i.e., a model that can generate data samples that look like the observed data), have been constructed in recent years based on \textit{adversarial}  learning \cite{goodfellow2014generative}.  The key idea of adversarial learning is to construct two separate networks, the generator and the discriminator, and have them compete against each other.  Succinctly, they have competing goals: the generator tries to randomly generate samples that the discriminator cannot tell apart from the real data, and the discriminator tries to classify whether it is a real sample (observed data) or fake sample (randomly generated).  A good generator will commonly fool the discriminator, and the samples will be qualitatively similar to the real samples. While generative models been commonly used in statistics and machine learning for a long time, recent advances in these adversarial techniques have made significant strides towards making photo-realistic fake imagery, text that practically reads like a human wrote it, and fake speech synthesis.  These advances are remarkable, but there is concern that the ability of such approaches could easily be damaging for society.  This could happen through ``fake news'' generation or evidence tampering.  They could also be made to artificially place individuals into videos (e.g., ``deepfakes''), and legal protections are ambiguous \cite{harris2019deepfakes}.  Currently, these systems are only good enough to trick someone in very narrow situations, such as generating pictures of hotel rooms \cite{radford2015unsupervised} or face generation \cite{karras2018progressive}. Generic systems that can generate any type of realistic image or text article are still far off, but are quickly improving.
 
In essence, machine learning methods can be very useful, but, as of this moment, they can do only what we train them to do, which is to recognize and repeat patterns in data.

\section*{Summary}    

Given the sheer amount of hype currently around machine learning, along with the staggering progress but also devastating failures, it has become hard to separate all this hype from the reality.

We wrote this document to help researchers navigate the current state of machine learning, and to understand why things have gone right, and where things have gone wrong. Looking over the last few years, problems have been caused by predictive models in the criminal justice system being proprietary (or generally uninterpretable), so that we cannot easily determine whether they are racially biased or miscalculated (see Section \ref{sec:interpretable}). Problems happen when proprietary models that are not well trained place people in danger (as given by the California wildfires example from Section \ref{sec:interpretable}). Related to the criminal justice situation, people often use variable importance of machine learning models to measure causal effects, when in fact this is not generally valid (Section \ref{sec:causal}). Machine learning methods have been known to be biased against groups of individuals when evaluating job applicant resumes \cite{hiringbias} when their designers do not take into account the way the algorithm would be used in practice when designing their models (Section \ref{sec:othercontraints}). Computer vision systems for self-driving cars can be fooled with adversarial examples because these adversarial examples are not natural images, like the data these systems were trained on. Black box algorithms for medical decisions often depend on confounders, such as the type of x-ray equipment used \cite{Zech2018} or leakage of data from the future, which would render them potentially useless in practice, and these confounders are extremely difficult to detect without an interpretable algorithm, and attempting to explain these black boxes is often not sufficient (Section \ref{sec:explain}). There are a vast number of people training black box neural networks when interpretable models would suffice that are much easier to train (Section \ref{sec:equally}). Very few people ever try to train an interpretable neural network (Section \ref{sec:equalnn}).

On the other hand, when done right, machine learning methods can have major successes that will lead to huge savings and efficiency throughout industry, government, and throughout society generally -- just imagine all of the failures of machine learning listed above being turned into successes.
We hope that by reading this TutORial, you will be sensitive to the potential pitfalls we illustrated above, and be more effective at high quality, successful, and impactful data science.

\bibliographystyle{TutORials}  
\bibliography{biblio.bib}                

\end{document}